\RequirePackage{letltxmacro}
\LetLtxMacro{\LaTeXtextbf}{\textbf}
\documentclass{ieeeaccess}
\LetLtxMacro{\textbf}{\LaTeXtextbf}
\usepackage{cite}
\usepackage{amsmath,amssymb,amsfonts}
\usepackage{algorithmic}
\usepackage{graphicx}
\usepackage{textcomp}

\makeatletter
\let\MYcaption\@makecaption
\makeatother

\usepackage[font=footnotesize]{subcaption}
\makeatletter
\let\@makecaption\MYcaption
\makeatother
\usepackage{subcaption}

\usepackage[linesnumbered,ruled,vlined]{algorithm2e}
\usepackage{multirow,booktabs,color,soul,threeparttable}
\usepackage[figuresright]{rotating}
\usepackage[table,xcdraw]{xcolor}

\DeclareMathOperator*{\argmax}{argmax} 
\DeclareMathOperator*{\argmin}{argmin}
\definecolor{hl}{rgb}{0.75,0.75,0.75}
\sethlcolor{hl}

\def\BibTeX{{\rm B\kern-.05em{\sc i\kern-.025em b}\kern-.08em T\kern-.1667em\lower.7ex\hbox{E}\kern-.125emX}}
\begin{document}
\history{Date of publication xxxx 00, 0000, date of current version xxxx 00, 0000.}
\doi{10.1109/ACCESS.2017.DOI}

\title{Reference Vector Adaptation and Mating Selection Strategy via Adaptive Resonance Theory-based Clustering for Many-objective Optimization}
\author{
	\uppercase{Takato Kinoshita}\authorrefmark{1},
	\uppercase{Naoki Masuyama}\authorrefmark{2}, \IEEEmembership{Member, IEEE},
	\uppercase{Yiping Liu}\authorrefmark{3}, \IEEEmembership{Member, IEEE},
	\uppercase{Yusuke Nojima}\authorrefmark{2}, \IEEEmembership{Member, IEEE},
	and \uppercase{Hisao Ishibuchi}\authorrefmark{4}, \IEEEmembership{Fellow, IEEE}.
}
\address[1]{Graduate School of Engineering, Osaka Prefecture University,
	1-1 Gakuen-cho Naka-ku, Sakai-Shi, Osaka 599-8531, Japan (e-mail: sbb01065@st.osakafu-u.ac.jp)}
\address[2]{Graduate School of Informatics, Osaka Metropolitan University,
	1-1 Gakuen-cho Naka-ku, Sakai-Shi, Osaka 599-8531, Japan (e-mails: \{masuyama,  nojima\}@omu.ac.jp)}
\address[3]{College of Computer Science and Electronic Engineering, Hunan University, Hunan 410082, China (e-mail: yiping0liu@gmail.com  )}
\address[4]{Guangdong Provincial Key Laboratory of Brain-inspired Intelligent Computation, Department of Computer Science and Engineering, Southern University of Science and Technology, Shenzhen 518055, China (e-mail: hisao@sustech.edu.cn)}
\tfootnote{This work was supported by Japan Society for the Promotion of Science (JSPS) KAKENHI Grant Number JP19K20358 and 22H03664. National Natural Science Foundation of China (Grant No. 62106073), the Natural Science Foundation of Hunan Province of China (Grant No. 2021JJ40116), the Fundamental Research Funds for the Central Universities (Grant No. HNU: 531118010537) National Natural Science Foundation of China (Grant No. 61876075), Guangdong Provincial Key Laboratory (Grant No. 2020B121201001), the Program for Guangdong Introducing Innovative and Enterpreneurial Teams (Grant No. 2017ZT07X386), The Stable Support Plan Program of Shenzhen Natural Science Fund (Grant No. 20200925174447003), Shenzhen Science and Technology Program (Grant No. KQTD2016112514355531)}

\markboth
{Author \headeretal: Preparation of Papers for IEEE TRANSACTIONS and JOURNALS}
{Author \headeretal: Preparation of Papers for IEEE TRANSACTIONS and JOURNALS}

\corresp{Corresponding author: Naoki Masuyama (e-mail: masuyama@omu.ac.jp).}

\begin{abstract}

Decomposition-based multiobjective evolutionary algorithms (MOEAs) with clustering-based reference vector adaptation show good optimization performance for many-objective optimization problems (MaOPs). Especially, algorithms that employ a clustering algorithm with a topological structure (i.e., a network composed of nodes and edges) show superior optimization performance to other MOEAs for MaOPs with irregular Pareto optimal fronts (PFs). These algorithms, however, do not effectively utilize information of the topological structure in the search process. Moreover, the clustering algorithms typically used in conventional studies have limited clustering performance, inhibiting the ability to extract useful information for the search process. This paper proposes an adaptive reference vector-guided evolutionary algorithm using an adaptive resonance theory-based clustering with a topological structure. The proposed algorithm utilizes the information of the topological structure not only for reference vector adaptation but also for mating selection. The proposed algorithm is compared with 8 state-of-the-art MOEAs on 78 test problems. Experimental results reveal the outstanding optimization performance of the proposed algorithm over the others on MaOPs with various properties.

\end{abstract}


\titlepgskip=-15pt

\maketitle

\section{Introduction}\label{sec:introduction}
\PARstart{M}{ost} optimization problems in the real world are categorized as Multiobjective Optimization Problems (MOPs).
They involve multiple objective functions to be optimized simultaneously.
In general, an MOP can be formulated as follows:
\begin{equation}\label{eq:MOP}
    \begin{aligned}
         & \text{min}
         &             & \boldsymbol{f}(\boldsymbol{x}) = (f_{1}(\boldsymbol{x}),\dots,f_{m}(\boldsymbol{x}),\dots,f_{M}(\boldsymbol{x})), \\
         & \text{s.t.}
         &             & \boldsymbol{x} \in S \subset \boldsymbol{R}^D,
    \end{aligned}
\end{equation}
where $\boldsymbol{x}$ is a $D$-dimensional decision vector in the search space $S$, $f_{m}$ is the $m$th objective function to be minimized ($m = 1, \dots, M$), and $M$ is the number of objectives.
Due to tradeoffs between the objective functions, no single solution usually optimizes all objective functions in an MOP simultaneously.
Alternatively, we obtain the Pareto optimal solution set (PS) and the Pareto optimal front (PF), which is the image of the PS in the objective space.

To solve MOPs, a large number of Multiobjective Evolutionary Algorithms (MOEAs) have been proposed over the last two decades.
The main advantage of MOEAs over other approaches to MOPs is that they can approximate the entire PF by their single run thanks to their population-based global search mechanisms. Thus, MOEAs are expected to have a strong convergence ability to push the population towards the PF as fast as possible as well as a strong diversification ability to cover the entire PF as wide as possible.

There are three categories of MOEAs, i.e., dominance-, indicator-, and decomposition-based MOEAs~\cite{DEA-GNG}.
The basic idea of the decomposition-based MOEAs is to decompose an MOP into a number of simple sub-problems and solve them simultaneously instead of the original one.
In conventional decomposition-based MOEAs, each single-objective problem is formed by a scalarizing function and a reference vector~\cite{MOEAD,RVEA}.
A reference vector specifies a sub-region in the objective space, and a localized single-objective optimization problem is generated by a scalarizing function and the reference vector.
Usually, reference vectors in the decomposition-based MOEAs are uniformly distributed over a simplex hyperplane in the entire objective space. Each reference vector $\boldsymbol{w} = \left(w_1, w_2, \dots, w_M\right)$ is generated so as to satisfy the following conditions:
\begin{equation}\label{eq:hyperplane}
    \sum_{m=1}^M w_m=1, \:
    w_{m} \in \left\{0,\frac{1}{H},\frac{2}{H},\dots,\frac{H}{H}\right\},
\end{equation}
where $H$ is an arrangement control parameter.
Finally, the decomposition-based MOEAs optimize each single-objective sub-problem while exchanging information with each other and output the union of all sub-problems' solutions as the solution set of the original MOP.

When an MOP has more than three objectives (i.e., $M>3$), it is often called a many-objective optimization problem (MaOP).
In recent years, MaOPs have been received increasing attention because of their challenging features~\cite{MaOP,RVEA-iGNG}.
For MaOPs, the performance of dominance-based MOEAs deteriorates due to the geometrical characteristics of high-dimensional space, i.e., almost all solutions in a population become non-dominated. 
On the other hand, the decomposition-based MOEAs can keep strong convergence pressure towards PF even when solving MaOPs~\cite{CEC2008_many}.
However, the decomposition-based MOEAs still have challenges in solving MaOPs in the real world because most of them have irregular PF shapes, e.g., disconnected, inverted, degenerated.
Some of the uniformly distributed vectors have no intersection with the PF if its shape is irregular.
Therefore, conventional decomposition-based MOEAs may acquire non-Pareto optimal and/or poorly distributed solutions corresponding to such vectors and diminish their searching efficiency~\cite{IshibuchiStudy,RVEA-iGNG}.

To tackle this issue, researchers have proposed various approaches.
One promising approach is adaptively adjusting the position of each reference vector based on the distribution of nodes generated by clustering algorithms which summarize information of obtained solutions in the search process.
In MOEA/D with Self Organized Map-based weight vectors (MOEA/D-SOM)~\cite{MOEAD-SOM}, SOM~\cite{SOM} is applied to MOEA/D, known as one of the representative decomposition-based MOEAs, and learns a network approximating the PF from the positions of solutions in the objective space.
In the Decomposition-based multiobjective Evolutionary Algorithm guided by Growing Neural Gas (DEA-GNG)~\cite{DEA-GNG}, GNG~\cite{GNG} is employed as a clustering algorithm instead of SOM.
GNG is suitable to handle irregular PFs because it converges more rapidly than SOM and requires no predefined topology in contrast with the requirement of SOM.
Liu, et al.~\cite{RVEA-iGNG} modify GNG in their Improved GNG (iGNG) to consider the population size in the learning process, and they adopt Reference Vector guided Evolutionary Algorithm (RVEA) as a base MOEA.
They name this algorithm RVEA-iGNG and demonstrate its high performance on various problems including irregular MaOPs and real-world problems.

Although it is common in such methodologies that we can discover more information from the approximated network than the node distribution, the approximated network is not yet utilized actively and effectively in the search process.
Moreover, clustering algorithms used in existing MOEAs are not always adequate for this purpose.

To address these issues, we introduce two improvements to the decomposition-based MOEAs using clustering algorithms; 1) a new searching mechanism utilizing additional information from the network, and 2) a highly stable clustering algorithm.
In this paper, we propose RVEA-CA as an improved algorithm of RVEA-iGNG by using the two improvements.
Specifically, as the first improvement, we propose a new mating selection strategy using the information of the topological structure (i.e., a network) obtained from a clustering algorithm.
As the second improvement, we apply Correntropy-induced metric-based Adaptive resonance theory (CA)~\cite{masuyama1} to RVEA-CA instead of iGNG.
CA is an Adaptive Resonance Theory (ART)~\cite{ART} based clustering algorithm, which measures a similarity between a training signal and existing nodes by using Correntropy-Induced Metric (CIM)~\cite{Correntropy}.
CIM is defined as a kernel function-based generalized similarity.
In addition, to cope with the change from iGNG to CA, we introduce an automatic parameter adjustment mechanism for CA to realize a similar functionality introduced in iGNG, which automatically adjusts the number of nodes to the population size.

The main contributions of this paper are as follows.

\begin{enumerate}
    \item
          We propose RVEA-CA as an improved algorithm of RVEA-iGNG.
          Both of them are based on RVEA and adjust the positions of reference vectors with a network generated by a clustering algorithm that summarizes information of obtained solutions in the search process.
          Their main difference can be summarized as follows: REVA-CA employs CA as a clustering algorithm instead of GNG and utilizes additional information of the network actively and effectively in the searching process.
          As a result, RVEA-CA achieves higher performance.

    \item
          To actively utilize additional information available from the network, we propose a new mating selection strategy that considers the network connectivity information.
          This strategy promotes the exploration of both known promising regions and potentially unexplored regions, maintaining a balance between the convergence and diversity in the search process.

    \item
          We introduce a mechanism to automatically adjust the parameters of CA based on the population size and solution distribution.
          While RVEA-iGNG achieves node number control by modifying the learning process of GNG, RVEA-CA achieves a similar functionality by just adaptively adjusting the parameters of CA without modifying the learning process of CA.
\end{enumerate}

The remainder of this paper is organized as below.
In Section \ref{sec:LiteratureReview}, we briefly discuss existing MOEAs as the motivation of the proposed algorithm.
In Section \ref{sec:Preliminals}, we introduce the basic algorithm of RVEA and CA.
In Section \ref{sec:ProposedMethod}, we explain the proposed algorithm.
In Section \ref{sec:Experiments}, we compare the search performance between the proposed algorithm and existing MOEAs.
Finally, some future research topics are suggested in Section \ref{sec:conclusion}.

\section{Literature Review}\label{sec:LiteratureReview}
\subsection{MOEAs with reference vector adaptation}\label{sec:Motivation}

When we solve an MOP, an intuitive choice is to use a dominance-based MOEA.
This is because the dominance relation, a kind of partial order, is valid in any MOPs, and it is always Pareto compliant.
In this point of view, Nondominated Sorting Genetic Algorithm III (NSGA-III)~\cite{NSGA-III-1} is one of the easy-to-use MOEAs for many-objective optimization.
NSGA-III is a dominance-based MOEA that employs a reference vector-based mechanism to maintain the diversity of solutions.
It controls the distribution of solutions by the distribution of the reference vectors.
There is also an adaptive variant for irregular PF shapes, A-NSGA-III~\cite{NSGA-III-1}.
A-NSGA-III generates reference vectors in the crowded region of solutions and deletes ones in the sparsely populated ones.

Vector angle-based Evolutionary Algorithm (VaEA)~\cite{VaEA} employs another approach.
Although VaEA is a dominance-based MOEA very similar to NSGA-III, it needs no reference vector.
Alternatively, it computes the angles of the dominated solutions from the non-dominated solution set and selects dominated solutions that have large angles to maintain the diversity.

Instead of using the dominance relation and a distribution criterion, we can choose other selection criteria derived from performance indicators.
Performance indicators are useful for solving MaOPs because we can use them to evaluate the contribution (i.e., fitness) of each solution as a scalar value  in the search process.
At least, they map the solutions in the objective space into the lower dimension feature space.
Some indicators require adapting a set of reference points to calculate the fitness appropriately.
Therefore, the indicator-based MOEAs which use such performance indicators also demand a mechanism to adjust reference points.
In this kind of algorithms, AR-MOEA~\cite{AR-MOEA} is promising.
AR-MOEA evaluates contributions of the solutions based on extended inverted generational distance (IGD)~\cite{IGD}.
In addition, it maintains an archive set to preserve non-dominated solutions that are promising in terms of extended IGD or have a high novelty.
Then, it adaptively updates the reference points by adding new points in the crowded region and deleting existing points in the sparse region based on the distribution of solutions in the archive set.

Of course, reference vector adaptation is also desired in decomposition-based MOEAs.
MOEA/D with Adaptive Weight Adjustment (MOEA/D-AWA)~\cite{MOEAD-AWA} is a decomposition-based MOEA extended the MOEA/D framework.
It employs an opposite approach of A-NSGA-III, which generates a new reference vector in a sparse region and deletes one in an overcrowded  region.
This approach is proposed based on geometrical analysis and its Pareto optimality is theoretically proven~\cite{MOEAD-AWA}.

AdaW~\cite{AdaW} is another example of decomposition-based MOEAs.
It considers not only the density but also the contribution.
It maintains an archive set to preserve well-distributed non-dominated solutions first, then generates reference vectors in unexplored and promising regions based on the distributions of solutions in the archive set, and deletes ones in crowded and unpromising regions.


\subsection{Reference vector adaptation by a clustering-based approach}

The basic idea of reference vector adaptation based on clustering algorithms is to control the distribution of reference vectors by using the distribution of nodes from clustering algorithms instead of using the density of solutions.
MOEA/D-SOM~\cite{MOEAD-SOM} is one of the earliest efforts to introduce a clustering algorithm into a decomposition-based MOEA.
It employs SOM that adapts the positions of nodes in the predefined network to the distribution of solutions. An archive set preserving solutions generated in the last several generations is input into SOM. 
SOM never changes the topology of the predefined network, i.e., it does not add or delete nodes or edges.
When we know that the PF shape is regular in advance, this may be a suitable property for MOEAs, because the population size of MOEAs is often fixed just like the number of nodes in the network learned by SOM.

\begin{figure*}[t]
    \centering
    \begin{minipage}[b]{\linewidth}
        \centering
        \includegraphics[width=0.2\linewidth]{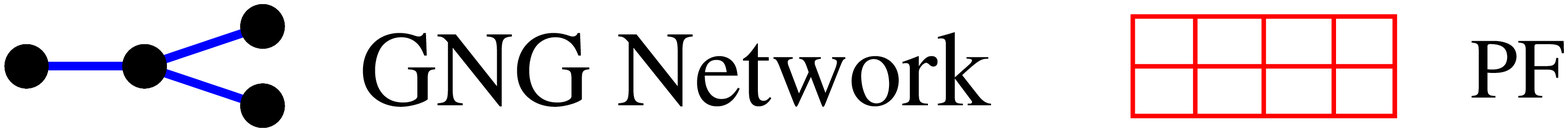}
    \end{minipage}\\
    \begin{minipage}[b]{0.32\linewidth}
        \centering
        \includegraphics[width=\linewidth]{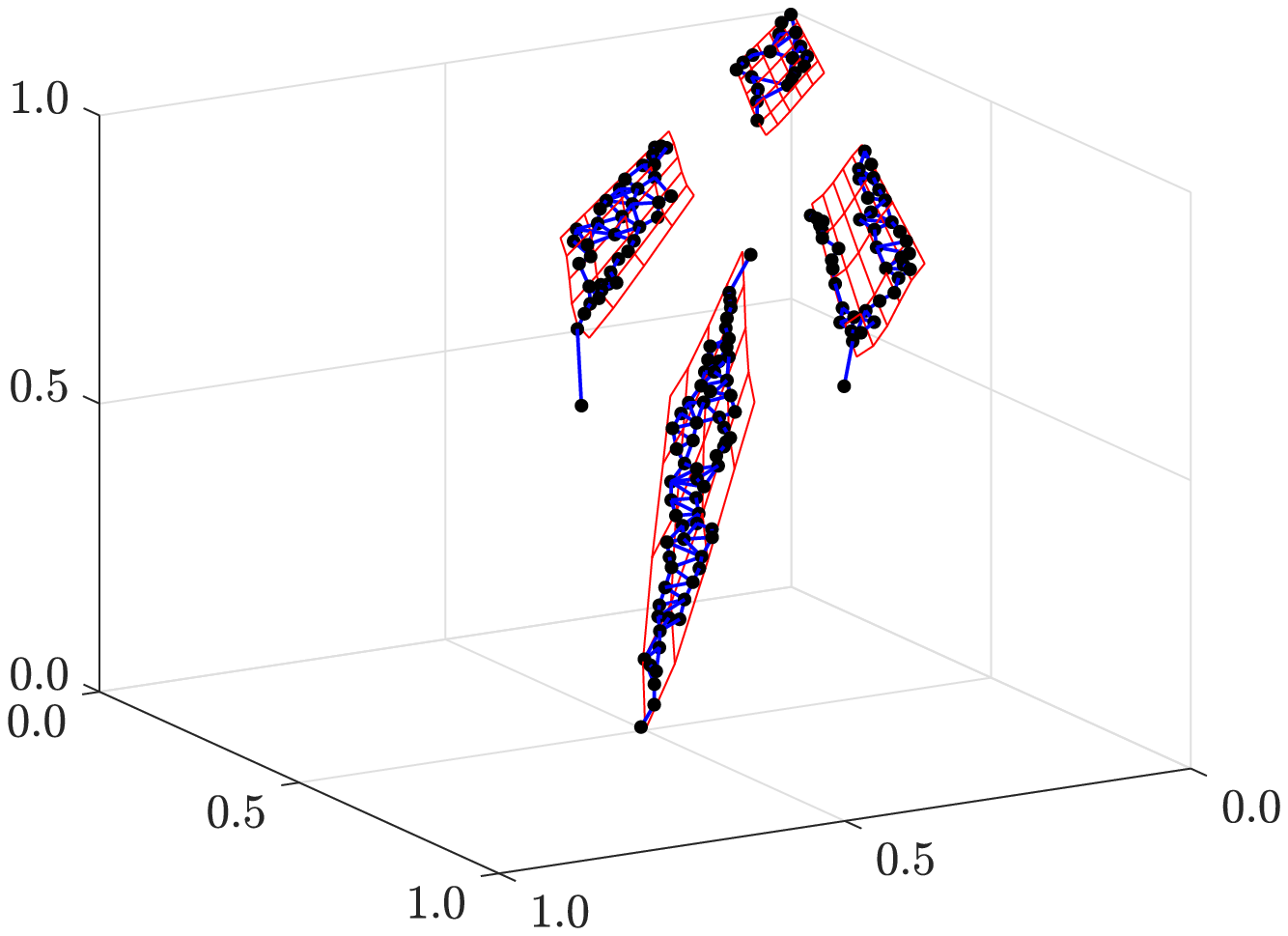}
        \subcaption{Best}\label{fig:best}
    \end{minipage}
    \begin{minipage}[b]{0.32\linewidth}
        \centering
        \includegraphics[width=\linewidth]{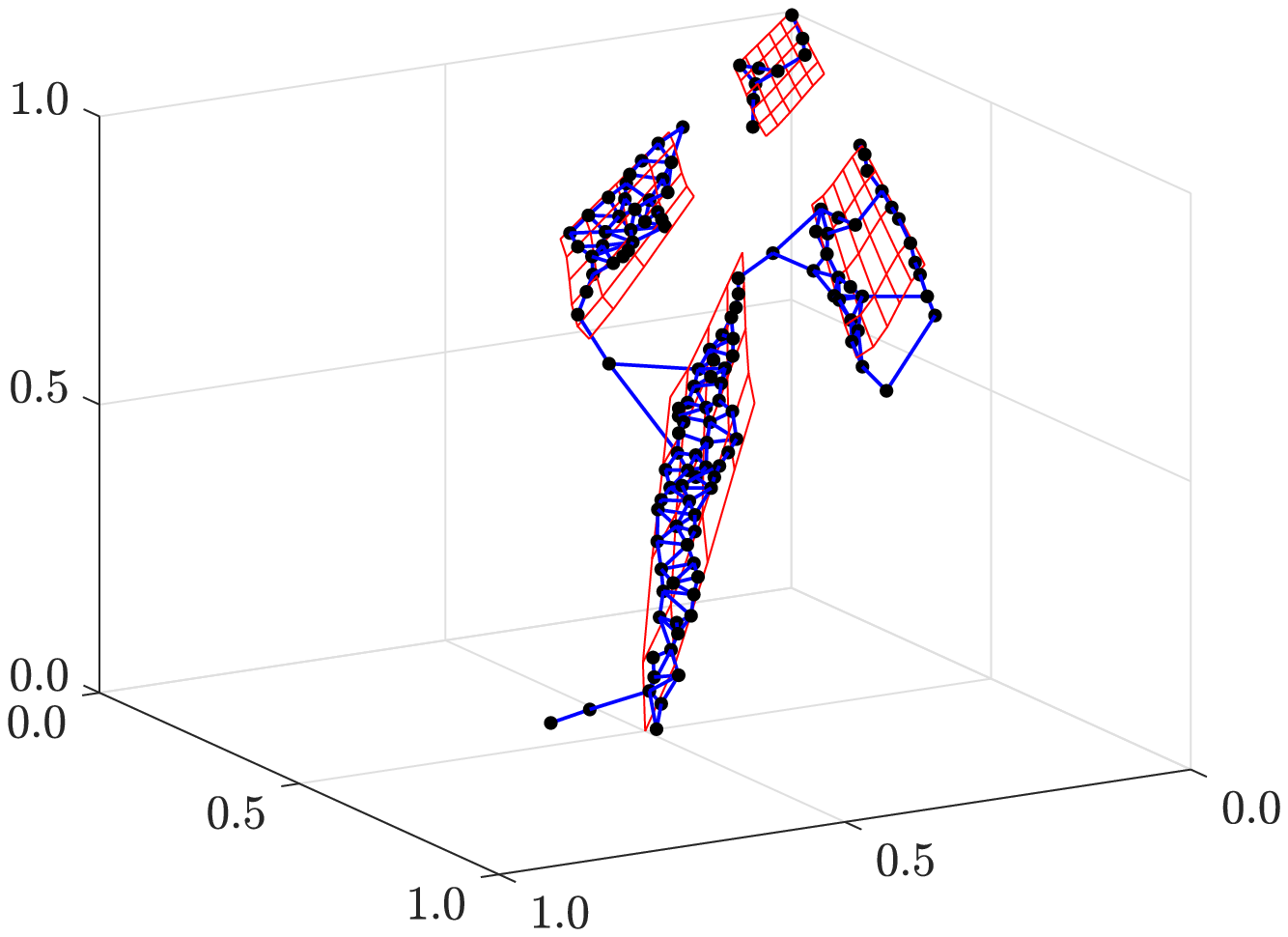}
        \subcaption{Median}\label{fig:median}
    \end{minipage}
    \begin{minipage}[b]{0.3\linewidth}
        \centering
        \includegraphics[width=\linewidth]{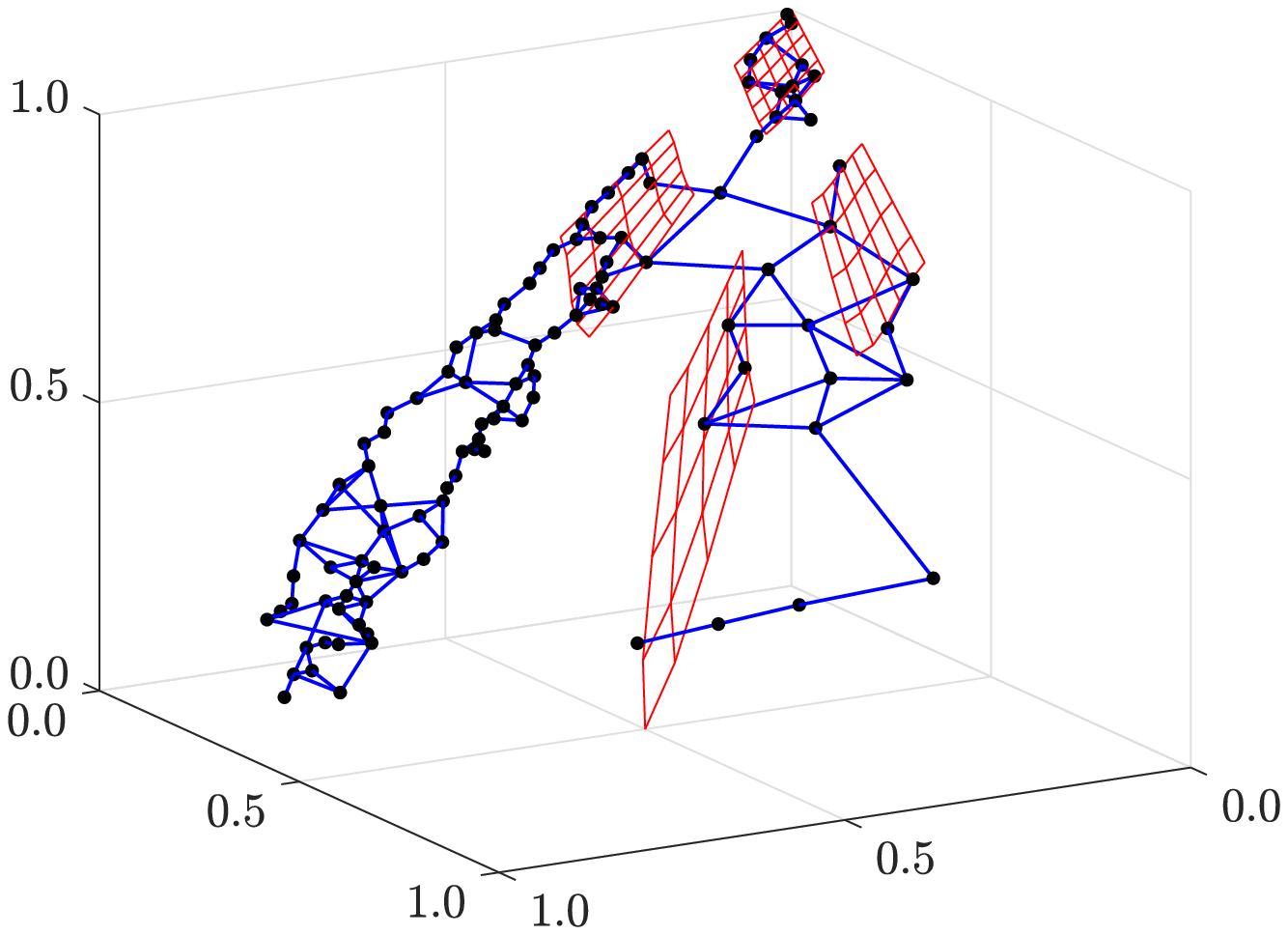}
        \subcaption{Worst}\label{fig:worst}
    \end{minipage}
    \caption{GNG network obtained from RVEA-iGNG on DTLZ7}\label{fig:iGNG}
\end{figure*}

However, for solving MOPs whose PF shapes are irregular, it is desirable that the clustering algorithm can adaptively change the topology of the network.
GNG is a clustering algorithm that satisfies this requirement.
GNG generates and adapts nodes and edges according to the series of inputs, and learns a network that depends on the distribution of the inputs in the dataset as a result.
DEA-GNG~\cite{DEA-GNG} employs GNG to handle irregular PFs and to adjust parameters of the scalarizing function.
It is a hybrid MOEA of dominance-based and decomposition-based MOEAs, and exploits GNG for the later mechanism.
In addition, it also maintains an archive set as a dataset to preserve promising solutions with respect to both dominance-based and decomposition-based mechanisms.

RVEA-iGNG~\cite{RVEA-iGNG} is a decomposition-base MOEA that improves the efficiency of GNG by modifying the learning process.
The canonical GNG removes only isolated nodes, i.e., nodes with no edges, as a noise reduction process.
However, it is possible that such a learning process is insufficient for the reference vector adaptation for MOEAs. More active node deletion in unpromising regions and addition in promising regions can improve the search performance.
iGNG, the improved GNG by RVEA-iGNG, deems nodes with fewer updates unnecessary, and then deletes them until the number of nodes is the same as the population size when the number of nodes exceeds 1.5 times the population size.
The archive set of RVEA-iGNG is maintained by $\epsilon$-indicator~\cite{UnaryEpsilonIndicator}, and its size is set to 2 times the population size.
This archive maintenance mechanism is superior in that the distribution of the existing reference vector set does not directly affect the selection criteria for solutions to be preserved.
In this point of view, RVEA-iGNG can be regarded as a hybrid MOEA of decomposition-based and indicator-based ones.
The most attractive point of RVEA-iGNG is its search performance.
In computational experiments~\cite{RVEA-iGNG}, it has demonstrated very competitive results in various MaOPs compared to other MOEAs, including categories other than decomposition-based.

Although decomposition-based MOEAs that employ reference vector adaptation by a clustering-based approach are promising, they do not actively and effectively utilize information other than node distribution.
For example, DEA-GNG utilizes network information to adjust the parameters of the scalarizing function, but it only plays an auxiliary role in the search process and does not utilize network information for operations other than the environmental selection~\cite{DEA-GNG}.

In the case of MOPs with disconnected PFs, the local region in each component of the PF can be regarded as a partial MOP of the original MOP.
Here, if the decomposition-based MOEAs can obtain a network with appropriate components from a clustering algorithm such as GNG, they can be expected to improve the search performance by utilizing the network information in the reference vector adaptation.
This idea is somewhat similar to that of MOEA/D-M2M~\cite{MOEAD-M2M}, i.e., decomposing a population not into individual solutions, but into some sub-populations.
However, SOM and GNG used in existing MOEAs are not suitable from this perspective.
SOM requires a predefined network topology, and the number of components cannot be changed adaptively during the search process.
iGNG is assumed to have learning instability.
Fig. \ref{fig:iGNG} shows the networks learned by iGNG during the search process of RVEA-iGNG on DTLZ7.
The search was performed 15 times independently, and the solution sets obtained in each run were compared to each other by hypervolume~\cite{HV}.
The network obtained in the best case is shown in Fig. \ref{fig:best}, the one in the median case in Fig. \ref{fig:median}, and the one in the worst case in Fig. \ref{fig:worst}.
In contrast to the best case, the worst case generates nodes and edges in locations that are significantly different from the true PF, indicating the learning instability of iGNG.
In addition, as seen in the median case, iGNG creates nodes in the middle of sub-networks, so it is not easy to get the appropriate topology of the PF from the network information.
iGNG also has a large number of hyperparameters, making it difficult to adaptively adjust them to the solution distribution or the PF shape.

\section{Preliminaries of RVEA and CA}\label{sec:Preliminals}


Our proposed RVEA-CA is based on RVEA and CA. This section explains RVEA and CA in detail. 

\subsection{RVEA}\label{sec:RVEA}

RVEA uses a reference vector set uniformly distributed on the hyperplane defined by (\ref{eq:hyperplane}).
Then, each solution $\boldsymbol{x}$ is associated with the reference vector that has the smallest angle $\theta$ with its objective function value $f(\boldsymbol{x})$, and environmental selection is performed in the corresponding sub-problem.
For any non-zero vectors $\boldsymbol{x}$ and $\boldsymbol{y}$, their angle $\theta$ is determined as follows:
\begin{equation}
    \theta=\operatorname{arccos} \frac{\langle \boldsymbol{x}, \boldsymbol{y}\rangle}{\|\boldsymbol{x}\| \|\boldsymbol{y}\|},
\end{equation}
where $\langle \boldsymbol{x}, \boldsymbol{y}\rangle$ is the inner product of $\boldsymbol{x}$ and $\boldsymbol{y}$, and $\|\boldsymbol{x}\|$ is the norm of $\boldsymbol{x}$.
Environmental selection in RVEA is based on the angle penalized distance (APD)~\cite{RVEA}, which is designed to maintain a balance between the convergence and diversity.
The APD of a solution $\boldsymbol{x}_{i}$ for a reference vector $\boldsymbol{w}_{j}$ is given
\begin{equation}\label{eq:APD}
    d_{ij}=\left(1+P\left(\theta_{i j}\right)\right) \cdot\left\|\boldsymbol{f}^{\prime}(\boldsymbol{x}_{i})\right\|,
\end{equation}
where $\boldsymbol{f}^{\prime}(\boldsymbol{x}_{i})$ is the normalized objective function vector of solution $\boldsymbol{x}_{i}$, and $\theta_{i j}$ is the angle between $\boldsymbol{f}^{\prime}(\boldsymbol{x}_{i})$ and $\boldsymbol{w}_{j}$.
$P(\cdot)$ denotes the penalty function for the angle given as follows:
\begin{equation}\label{eq:Penalty}
    P\left(\theta_{ij}\right)=M \cdot\left(\frac{t}{t_{\max }}\right)^{\alpha} \cdot \frac{\theta_{ij}}{\gamma_{j}},
\end{equation}
where $M$ is the number of objectives, $t$ is the number of generations, and $\gamma_{j}$ is the minimum angle between the reference vector $\boldsymbol{w}_{j}$ and the other reference vectors, which is used to normalize $\theta_{ij}$.
The smaller the APD value, the more likely the solution is to be selected.
APD-based environmental selection can promote both the convergence and diversity simultaneously.
However, the APD function controls the balance between the convergence and diversity with the hyperparameter $\alpha$ so that convergence is prioritized in early generations and the diversity in later generations.
APD can be regarded as an angle-based scalarizing function that can be normalized easier than Euclidean distance-based scalarizing functions~\cite{RVEA}.

\subsection{CA}\label{sec:CA}

ART-based clustering algorithms have been shown to be more stable and to have faster self-organizing performance than GNG-based algorithms \cite{ASOINN, SOINNP} while suppressing excessive node creation.
Among ART-based clustering algorithms, CA is an easy-to-use algorithm because it has a smaller number of hyperparameters than the others~\cite{masuyama2, masuyama1, masuyama3}.

In this section, we first briefly present the theoretical background of CIM and then describe the overall learning process of CA.

\subsubsection{Correntropy and Correntropy-Induced Metric}\label{sec:CIM}

Correntropy~\cite{Correntropy} $C(\boldsymbol{x},\boldsymbol{y})$ provides a generalized similarity measure between two arbitrary instances $\boldsymbol{x}=(x_{1},x_{2},\dots,x_{d})$ and $\boldsymbol{y}=(y_{1},y_{2},\dots,y_{d})$ as follows:
\begin{equation}\label{eq:Correntropy}
    C(\boldsymbol{x}, \boldsymbol{y})=\mathbf{E}\left[\kappa_{\sigma}(\boldsymbol{x}-\boldsymbol{y})\right],
\end{equation}
where $\mathbf{E}[\cdot]$ is the expectation operation, and $\kappa_{\sigma}(\cdot)$ denotes a positive definite kernel with a kernel bandwidth $\sigma$.
In general, the dimension of an instance is finite.
Thus, the estimator for the correntropy $C(\boldsymbol{x},\boldsymbol{y})$ can be defined as follows:
\begin{equation}\label{eq:estCorrentropy}
    \hat{C}(\boldsymbol{x}, \boldsymbol{y})=\frac{1}{d} \sum_{i=1}^{d} \kappa_{\sigma}\left(x_{i}-y_{i}\right).
\end{equation}

In this paper, we use the following Gaussian kernel in the correntropy:
\begin{equation}\label{eq:Gaussian}
    \kappa_{\sigma}\left(x_{i}-y_{i}\right)=\exp \left[-\frac{\left(x_{i}-y_{i}\right)^{2}}{2 \sigma^{2}}\right],
\end{equation}
where $\sigma$ is a kernel bandwidth.

A nonlinear metric called CIM is derived from the correntropy.
The CIM quantifies the similarity between two instances as follows:
\begin{equation}\label{eq:CIM}
    \operatorname{CIM}(\boldsymbol{x}, \boldsymbol{y}, \sigma)=\left[\kappa_{\sigma}(0)-\hat{C}(\boldsymbol{x}, \boldsymbol{y})\right]^{\frac{1}{2}},
\end{equation}
where $\kappa_{\sigma}(0)=1$ from (\ref{eq:Gaussian}).
Here, thanks to the Gaussian kernel without a coefficient $\frac{1}{\sqrt{2\pi}\sigma}$ as defined in (\ref{eq:Gaussian}), a range of the CIM is limited to $[0,1]$.

\subsubsection{Learning Procedure}\label{sec:CALearning}

We use the following notations: Training instances are $\boldsymbol{X}=\left(\boldsymbol{x}_{1},\boldsymbol{x}_{2},\dots,\boldsymbol{x}_{N_{X}}\right)\ \left(N_{X}\in\mathbb{Z}^{+}\right)$ where $\boldsymbol{x}_{n} = \left(x_{n1},x_{n2},\dots,x_{nd}\right)$ is a $d$-dimensional feature vector.
A collection of prototype nodes in CA at the point of the presentation of instance $\boldsymbol{x}_{n}$ is $\boldsymbol{Y} = \left(\boldsymbol{y}_{1},\boldsymbol{y}_{2},\dots,\boldsymbol{y}_{K}\right)\ \left(K\in\mathbb{Z}^{+}\right)$ where a node $\boldsymbol{y}_{k} = \left(y_{k1},y_{k2},\dots,y_{kd}\right)$ has the same dimension as $\boldsymbol{x}_{n}$.
The edge set in CA is $E$, a non-directed edge $e_{\{k,l\}} = e_{\{l,k\}}$ joins node $\boldsymbol{y}_{k}$ and node $\boldsymbol{y}_{l}$, and each edge has an age, i.e., given the age function $\operatorname{age}$ from $E$ to $\mathbb{R}^{+}$.
Furthermore, each node $\boldsymbol{y}_{k}$ has an individual bandwidth $\sigma$ for the CIM, i.e., $\boldsymbol{S} = \left(\sigma_{1},\sigma_{2},\dots,\sigma_{K}\right)$ and has the neighborhood, i.e., $N(\boldsymbol{y}_{k})=\left\{l \mid e_{\{k,l\}} \in E\right\}$.

The learning procedure of CA is summarized in Algorithm \ref{alg:TrainCA}.
It is divided into four parts: 1) initialization process for a bandwidth of a kernel function in the CIM, 2) winner node selection, 3) vigilance test, and 4) node and edge learning.
Each of them is explained in the follows.

\textbf{1. Initialization Process for a Bandwidth of a Kernel Function:}
Similarity measurement between an instance and a node has a large impact on the performance of clustering algorithms.
CA uses the CIM as a similarity measure.
As defined in (\ref{eq:CIM}), the state of the CIM is controlled by a bandwidth $\sigma$ of a kernel function which is a data-dependent parameter.

In general, the bandwidth of a kernel function can be estimated from $\lambda$ instances belonging to a certain distribution, which is defined as follows:
\begin{gather}
    \boldsymbol{\Sigma} = U(F_{\nu})\boldsymbol{\Gamma}\lambda^{\frac{1}{2\mu+d}}, \label{eq:Sigma}\\
    U\left(F_{\nu}\right)=\left(\frac{\pi^{d / 2} 2^{d+\nu-1}(\nu !)^{2} R(F)^{d}}{\nu \kappa_{\nu}^{2}(F)\left[(2 \nu) ! !+(d-1)(\nu ! !)^{2}\right]}\right)^{\frac{1}{2 \nu+d}}, \label{eq:U}
\end{gather}
where $\boldsymbol{\Gamma}$ denotes a rescale operator ($d$-dimensional vector) which is defined by a standard deviation of the $d$ attributes among $\lambda$ instances.
$\nu$ is the order of a kernel.
$R(F)$ is a roughness function.
$\kappa_{\nu}(F)$ is the moment of a kernel.
In this paper, we utilize the Gaussian kernel for the CIM.
Therefore, $\nu = 2$, $R(F) = \left(2\sqrt{\pi}\right)^{-1}$, and $\kappa_{\nu}^{2}(F) = 1$ are derived.
The details of the derivation of (\ref{eq:Sigma}) and (\ref{eq:U}) can be found in \cite{henderson12}.

\begin{algorithm}[t]
    \DontPrintSemicolon
    \caption{Learning procedure of CA}
    \label{alg:TrainCA}
    \KwIn{\\
        \begin{itemize}
            \setlength{\itemsep}{-0.6mm}
            \item[-] training instances: $ \boldsymbol{X} = \left( \boldsymbol{x}_{1}, \boldsymbol{x}_{2},\ldots, \boldsymbol{x}_{N_{X}} \right) $ ($ N_{X} \in \mathbb{Z}^{+} $), \\
            \item[-] prototype graph: $G = (\boldsymbol{Y}, E)$; $ \boldsymbol{Y} = \left(\boldsymbol{y}_{1}, \boldsymbol{y}_{2}, \ldots, \boldsymbol{y}_{K}\right) $ ($ K \in \mathbb{Z}^{+} $), \\
            \item[-] kernel bandwidths of $ \boldsymbol{Y} $: $ \boldsymbol{S} = \left( \sigma_{1},\sigma_{2},\ldots, \sigma_{K}\right)  $, \\
            \item[-] winner counts of $ \boldsymbol{Y} $: $ \boldsymbol{A} = (\alpha_{1}, \alpha_{2}, \ldots, \alpha_{K}) $, \\
            \item[-] an interval for adapting $ \sigma $: $ \lambda $, \\
            \item[-] and a similarity threshold: $ V $. \\
        \end{itemize}
    }
    \KwOut{\\
        \vspace{-0.3\baselineskip}
        \begin{itemize}
            \item[-] updated graph: $ G = (\boldsymbol{Y}, E) $. \\
        \end{itemize}
    }
    \SetKwBlock{Begin}{function}{end function}
    \Begin( \text{TrainCA($ \boldsymbol{X} $, $ G = (\boldsymbol{Y}, E) $, $ \boldsymbol{S} $, $ \boldsymbol{A} $, $ \lambda $, $ V $)}){
    Input an instance $ \boldsymbol{x}_{n} $. \\
    \If{\upshape The instant $ n $ is multiple of $ \lambda $ }{
        Compute a kernel bandwidth for the CIM by (\ref{eq:sigma}).
    }
    \uIf{ $ K < 1 $ }{
        Generate a new node as $ \boldsymbol{y}_{K+1} \gets \boldsymbol{x}_{n} $. \\
        Assign a kernel bandwidth $ \sigma_{K+1} $. \\
        Initialize winner counter $ \alpha_{K+1} \gets 1 $.
    }\Else{
    Compute the CIM by (\ref{eq:CIM}). \\
    Search indexes of winner nodes $ k_{1} $ and $ k_{2} $ by (\ref{eq:winner1}) and (\ref{eq:winner2}), respectively. \\
    \uIf{ \upshape$ V < \mathrm{CIM}\left(\boldsymbol{x}_{n}, \boldsymbol{y}_{k_{1}}, \mathrm{mean}(S) \right) $ }{
        Generate a new node as $ \boldsymbol{y}_{K+1} \gets \boldsymbol{x}_{n} $. \\
        Compute a kernel bandwidth $ \sigma_{K+1} $ which is defined by (\ref{eq:sigma}). \\
        Initialize winner counter $ \alpha_{K+1} \gets 1 $. \\
    }\Else{
    Update all edges connecting to $\boldsymbol{y}_{k_{1}}$ by (\ref{eq:updateAge}). \\
    Update a state of $ \boldsymbol{y}_{k_{1}} $ by (\ref{eq:updateK1}). \\
    Update $ \alpha_{k_{1}} $ by (\ref{eq:updateAlpha}). \\
    \If{ \upshape$ V \geq \mathrm{CIM}\left(\boldsymbol{x}_{n}, \boldsymbol{y}_{k_{2}}, \mathrm{mean}(S) \right) $ }{
    Update a state of $ \boldsymbol{y}_{k_{2}} $ by (\ref{eq:updateK2}). \\
    Search the most similar node $\boldsymbol{y}_{l}$ to $\boldsymbol{y}_{k_{1}}$ in $N\left(\boldsymbol{y}_{k_{1}}\right)$. \\
    \If{ \upshape$\operatorname{age}\left(e_{\{k_{1},l\}}\right) > \left|N\left(\boldsymbol{y}_{k_{1}}\right)\right|$ }{
        Update $E$ by (\ref{eq:delteEdge}). \\
    }
    }
    Update $E$ by (\ref{eq:updateE}). \\
    }
    }
    \If{$ n \leq N_{X} $}{
        Continue from step 2 with $ n \gets n+1 $. \\
    }
    \Return{ \upshape $ G = (\boldsymbol{Y},E) $}. \\
    }
\end{algorithm}

In CA, the initial state of $\sigma$ in the CIM is defined by training instances.
When a new node $\boldsymbol{y}_{K}$ is generated from $\boldsymbol{x}_{n}$, a bandwidth $\sigma_{K+1}$ is estimated from the past $\lambda$ instances, i.e., $\left(\boldsymbol{x}_{n-\lambda},\dots,\boldsymbol{x}_{n-2},\boldsymbol{x}_{n-1}\right)$, by using (\ref{eq:Sigma}) and (\ref{eq:U}) with $\nu = 2$, $R(F) = \left(2\sqrt{\pi}\right)^{-1}$, and $\kappa_{\nu}^{2}(F) = 1$, as follows:
\begin{equation}\label{eq:estSigma}
    \boldsymbol{\Sigma}=\left(\frac{4}{2+d}\right)^{\frac{1}{4+d}} \boldsymbol{\Gamma} \lambda^{-\frac{1}{4+d}},
\end{equation}
where $\boldsymbol{\Gamma}$ denotes a rescale operator ($d$-dimensional vector) which is defined by a standard deviation of the $d$ attributes among $\lambda$ instances of CA.
Here, $\boldsymbol{\Sigma}$ contains the bandwidth of each attribute.
In this paper, the median of $\boldsymbol{\Sigma}$ is selected as a representative bandwidth for the new node $\boldsymbol{y}_{K+1}$, i.e.,
\begin{equation}\label{eq:sigma}
    \sigma_{K+1}=\operatorname{median}(\boldsymbol{\Sigma}).
\end{equation}

\textbf{2. Winner Node Selection:}
Once an instance $\boldsymbol{x}_{n}$ is presented to CA, two nodes which have a similar state to the instance $\boldsymbol{x}_{n}$ are selected, namely, winner nodes $\boldsymbol{y}_{k_{1}}$ and $\boldsymbol{y}_{k_{2}}$.
The winner nodes are determined based on the state of the CIM as follows:
\begin{gather}
    k_{1}=\argmin _{1 \le k \le K}\left[\operatorname{CIM}\left(\boldsymbol{x}_{n}, \boldsymbol{y}_{k}, \operatorname{mean}(\boldsymbol{S})\right)\right], \label{eq:winner1}\\
    k_{2}=\argmin_{k \neq k_{1}}\left[\operatorname{CIM}\left(\boldsymbol{x}_{n}, \boldsymbol{y}_{k}, \operatorname{mean}(\boldsymbol{S})\right)\right], \label{eq:winner2}
\end{gather}
where $k_{1}$ and $k_{2}$ denote indexes of the 1st and 2nd winner nodes i.e., $\boldsymbol{y}_{k_{1}}$ and $\boldsymbol{y}_{k_{2}}$, respectively.
$\boldsymbol{S}$ is a bandwidth for a kernel function of the CIM in each node.

Note that when there is no node in CA, the $(\lambda+1)$th instance becomes the initial node (i.e., $\boldsymbol{y}_{1} \gets \boldsymbol{x}_{\lambda+1}$).
In this case, the bandwidth of $\boldsymbol{y}_{1}$ is estimated from the 1st to $\lambda$th instances in a collection of training instances $\boldsymbol{X}$ by (\ref{eq:Sigma})-(\ref{eq:sigma}), and the next instance is given without vigilance test until the 1st and 2nd winner nodes can be defined.

\textbf{3. Vigilance Test:}
Similarities between an instance $\boldsymbol{x}_{n}$ and the 1st and 2nd winner nodes are defined as follows:
\begin{gather}
    V_{k_{1}}=\operatorname{CIM}\left(\boldsymbol{x}_{n}, \boldsymbol{y}_{k_{1}}, \operatorname{mean}(\boldsymbol{S})\right), \label{eq:simWinner1}\\
    V_{k_{2}}=\operatorname{CIM}\left(\boldsymbol{x}_{n}, \boldsymbol{y}_{k_{2}}, \operatorname{mean}(\boldsymbol{S})\right). \label{eq:simWinner2}
\end{gather}

The vigilance test classifies the relationship between an instance and a node into three cases by using a predefined similarity threshold $V$.

\begin{itemize}
    [
        \setlength{\IEEElabelindent}{\dimexpr-\labelwidth-\labelsep}
        \setlength{\itemindent}{\dimexpr\labelwidth+\labelsep}
        \setlength{\listparindent}{\parindent}
    ]
    \item Case I
\end{itemize}

The similarity between an instance $\boldsymbol{x}_{n}$ and the 1st winner node $\boldsymbol{y}_{k_{1}}$ is larger (i.e., less similar) than the similarity threshold $V$, namely:
\begin{equation}\label{eq:cond1}
    V < V_{k_{1}} \le V_{k_{2}}.
\end{equation}

If (\ref{eq:cond1}) is satisfied, $V_{k_{2}} > V$ is also satisfied since $V_{k_{2}} \le V_{k_{1}} > V$.
Thus, a new node is defined as $\boldsymbol{y}_{K+1} \gets \boldsymbol{x}_{n}$, and the bandwidth $\sigma_{K+1}$ is defined by (\ref{eq:sigma}).

Moreover, the winning counter $\alpha_{K+1}$ is initialized.
$\alpha_{K+1}$ is the number of instances that have been accumulated by a node, which is initialized as $\alpha_{K+1} \gets 1$.

\begin{itemize}
    [
        \setlength{\IEEElabelindent}{\dimexpr-\labelwidth-\labelsep}
        \setlength{\itemindent}{\dimexpr\labelwidth+\labelsep}
        \setlength{\listparindent}{\parindent}
    ]
    \item Case II
\end{itemize}

The similarity between an instance $\boldsymbol{x}_{n}$ and the 1st winner node $\boldsymbol{y}_{k_{1}}$ is smaller (i.e., more similar) than the similarity threshold $V$, and the similarity between the instance $\boldsymbol{x}_{n}$ and the 2nd winner node $\boldsymbol{y}_{k_{2}}$ is larger (i.e., less similar) than the similarity threshold $V$, namely:
\begin{equation}\label{eq:cond2}
    V_{k_{1}} \le V < V_{k_{2}}.
\end{equation}

If (\ref{eq:cond2}) is satisfied, the winning counter $\alpha_{k_{1}}$ is updated first as follows:
\begin{equation}\label{eq:updateAlpha}
    \alpha_{k_{1}}\gets\alpha_{k_{1}}+1.
\end{equation}

Then, node and edge learning is performed.
In addition, a new edge between the 1st winner node and the 2nd winner node is defined as $\operatorname{age}\left(e_{\{k_{1},k\}}\right) \gets 0$, and the edge set is updated as follows:
\begin{equation}\label{eq:updateE}
    E \gets E \cup \left\{e_{\{k_{1},k_{2}\}}\right\}.
\end{equation}

\begin{itemize}
    [
        \setlength{\IEEElabelindent}{\dimexpr-\labelwidth-\labelsep}
        \setlength{\itemindent}{\dimexpr\labelwidth+\labelsep}
        \setlength{\listparindent}{\parindent}
    ]
    \item Case III
\end{itemize}

Similarities between an instance $\boldsymbol{x}_{n}$ and the 1st and 2nd winner nodes are both in range of the similarity threshold $V$, namely:
\begin{equation}\label{eq:cond3}
    V_{k_{1}} \le  V_{k_{2}} \le V.
\end{equation}

If (\ref{eq:cond3}) is satisfied, the winning counter $\alpha_{k_{1}}$ is updated by (\ref{eq:updateAlpha}), and the node and edge learning is performed.

\textbf{4. Node and Edge Learning:}
First, the age of each edge connected to the 1st winner node, i.e., $e_{\{k_{1},k\}} \in E$, is updated as follows:
\begin{equation}\label{eq:updateAge}
    \operatorname{age}\left(e_{\{k_{1},k\}}\right) \gets \operatorname{age}\left(e_{\{k_{1},k\}}\right) + \frac{1}{\left|N\left(\boldsymbol{y}_{k_{1}}\right)\right|}.
\end{equation}

Then, different node and edge learning is performed based on the results of the vigilance test.

If Case II, the state of the 1st winner node $\boldsymbol{y}_{k_{1}}$ is updated as follows:
\begin{equation}\label{eq:updateK1}
    \boldsymbol{y}_{k_{1}} \gets \boldsymbol{y}_{k_{1}}+\frac{1}{\alpha_{k_{1}}}\left(\boldsymbol{x}_{n}-\boldsymbol{y}_{k_{1}}\right).
\end{equation}

When updating the node, the amount of change is divided by $\alpha_{k_{1}}$. Thus, the larger $\alpha_{k_{1}}$ becomes, the smaller the node position changes.

If Case III, the state of the 1st winner node $\boldsymbol{y}_{k_{1}}$ is updated by (\ref{eq:updateK1}) and the state of the 2nd winner node $\boldsymbol{y}_{k_{2}}$ is updated as follows:
\begin{equation}\label{eq:updateK2}
    \boldsymbol{y}_{k_{2}} \gets \boldsymbol{y}_{k_{2}}+\frac{1}{10\alpha_{k_{2}}}\left(\boldsymbol{x}_{n}-\boldsymbol{y}_{k_{2}}\right).
\end{equation}

In addition, the edge deleting procedure is performed.
First, the node $\boldsymbol{y}_{l}$ that has the largest similarity (i.e., worst similar) between the 1st winner node $\boldsymbol{y}_{k_{1}}$ in the neighborhood of $\boldsymbol{y}_{k_{1}}$ is searched as follows:
\begin{equation}\label{eq:findDel}
    l = \argmax_{k \in N\left(\boldsymbol{y}_{k_{1}}\right)}\left[\operatorname{CIM}\left(\boldsymbol{y}_{k}, \boldsymbol{y}_{k_{1}}, \operatorname{mean}(\boldsymbol{S})\right)\right].
\end{equation}

Then, if $\operatorname{age}\left(e_{\{k_{1},l\}}\right) > \left|N\left(\boldsymbol{y}_{k_{1}}\right)\right|$ is satisfied, the edge between $\boldsymbol{y}_{k_{1}}$ and $\boldsymbol{y}_{l}$ is deleted, and the edge set $E$ is updated as follows:
\begin{equation}\label{eq:delteEdge}
    E \gets E \setminus \left\{e_{\{k_{1},l\}}\right\}.
\end{equation}



\section{Proposed Algorithm: RVEA-CA}\label{sec:ProposedMethod}

The main idea of RVEA-CA is to replace the clustering algorithm in RVEA-iGNG with CA from GNG.
Like RVEA-iGNG, RVEA-CA uses a reference vector set whose distribution is adaptively determined from the distribution of nodes in the network generated by a clustering algorithm, instead of using the fixed and uniformly distributed reference vector set of RVEA.
In addition, the proposed algorithm employs a new mating selection strategy based on the network information.
It utilizes information obtained in the search process for both reproduction and selection of solutions and achieves both an intensive search to known local regions and a search to promising unexplored regions.

\subsection{General Framework of RVEA-CA}

\begin{algorithm}[t!]
    \DontPrintSemicolon
    \caption{\newline General Framework of RVEA-CA}
    \label{alg:RVEA-CA}
    \KwIn{
        \begin{itemize}
            \setlength{\itemsep}{-0.6mm}
            \item[-] objective function: $\boldsymbol{f}$,
            \item[-] initial population: $P$,
            \item[-] population size: $N$,
            \item[-] maximum generations: $t_{\max}$,
            \item[-] and an interval for adapting $ \sigma $: $ \lambda $.
        \end{itemize}
    }
    \KwOut{
        \begin{itemize}
            \setlength{\itemsep}{-0.6mm}
            \item[-] final population: $P$,
            \item[-] and objective values of $P$: $\boldsymbol{f}[P]$.
        \end{itemize}
    }
    \vspace{2mm}
    \SetKwBlock{Begin}{function}{end function}
    \Begin(\text{RVEA-CA($\boldsymbol{f}$, $P$, $N$, $t_{\max}$, $ \lambda $)}){
    $t \gets 1$\\
    $V \gets 0.1$; $\boldsymbol{Y} \gets \emptyset$; $E \gets \emptyset$; $G \gets (\boldsymbol{Y},E)$; $P_{A} \gets \emptyset$\\
    $\boldsymbol{z}^{\min} \gets \min\left(\boldsymbol{f}\left[P\right]\right)$\\
    $P_{A} \gets \operatorname*{UpdateArchive}_{\boldsymbol{f}}(P,P_{A},N)$\\
    \While{$t<t_{\max}$}{
    $O \gets \operatorname*{ClusterBasedReproduction}_{\boldsymbol{f}}(P,G,\boldsymbol{z}^{\min})$\\
    $\boldsymbol{z}^{\min} \gets \min\left(\left\{\boldsymbol{z}^{\min}\right\} \cup \boldsymbol{f}\left[O\right]\right)$ \\
    $P \gets P \cup O$ \\
    $P_{A} \gets \operatorname*{UpdateArchive}_{\boldsymbol{f}}(P,P_{A},N)$ \\
    $P \gets P\cup P_{A}$ \\
    \If{$t \leq 0.9t_{\max}$}{
        $X \gets \operatorname{MapToHyperplane}(\boldsymbol{f}[P],\boldsymbol{z}^{\min})$\\
        $\boldsymbol{X} \gets \operatorname{RandIndex}(X)$\\
        $\boldsymbol{Y},\, E,\, V \gets \operatorname{AdaptiveTrainCA}(\boldsymbol{X},\lambda,V,N)$ \\
        $G \gets (\boldsymbol{Y},E)$\\
        $R \gets \boldsymbol{Y}$\\
        \If{$t = 0.9t_{\max}$}{
            $Y^{\prime} \gets \operatorname{MapToHyperplane}(\boldsymbol{f}[P_{A}],\boldsymbol{z}^{\min})$ \\
            $R \gets R \cup Y^{\prime}$ \\
        }
    }
    $P\gets\operatorname*{ESelection}_{\boldsymbol{f}}(P,R,t,t_{\max},N,\boldsymbol{z}^{\min})$ \\
    $t \gets t + 1$ \\
    }
    \Return{ \upshape $P$, and $\boldsymbol{f}\left[P\right]$}.
    }
\end{algorithm}

The overall framework of RVEA-CA is shown in Algorithm \ref{alg:RVEA-CA}.
It is divided into the following four parts:

\begin{enumerate}
    \item
          Initialize (Lines 2-4):
          RVEA-CA initializes the archive set, estimates the ideal point $\boldsymbol{z}^{\min}$ based on the initial population, and initializes the similarity threshold $V$ with a small value.
          In this paper, the initial value of $V$ is set to 0.1.
          It also initializes the node collection $\boldsymbol{Y}$ and the edge set $E$.
          In the proposed algorithm we do not specify the method of generating the initial population, but in this paper we use a randomly generated population.
          $\boldsymbol{z}^{\min}$ is defined by the minimum value of each objective function in all solutions.

    \item
          Reproduction and Update Archive (Lines 5-11):
          RVEA-CA generates an offspring population from the current population, and then the archive set is updated with the union of the two sets.
          By employing the mating selection strategy based on clustering, the proposed algorithm generates offspring solutions in both promising and unexplored regions in an MOP.
          In the proposed algorithm, crossover and mutation operators are not specified.
          As in RVEA-iGNG, the archive set is managed by the $\varepsilon$-indicator every generation so that more promising solutions are retained in the archive set.
          Moreover, $\boldsymbol{z}^{\min}$ is updated by the offspring solutions.

    \item
          Training CA (Lines 13-16):
          Like RVEA-iGNG, RVEA-CA uses the union of the population and the archive set as training data and generates a reference vector set by clustering the training data in each generation.
          However, while RVEA-iGNG updates the GNG network inherited from the previous generation, RVEA-CA inherits only the similarity threshold $V$ and learns from a state without nodes in each generation.
          As in RVEA-iGNG, in the last 10\% of all generations, the reference vector set is fixed, and the archive set is also added to it.
          Note that since the population and the archive set added to the reference vector set are mapped onto the hyperplane, the nodes and the reference vector set are also located on the same hyperplane.

    \item
          Environmental Selection (Lines 17-21):
          RVEA-CA selects solutions with the reference vector set and the APD function.
          Since the number of reference vectors is not necessarily the same as the population size, the additional operation based on novelty in the objective space is performed to compensate for the excess or shortage of solutions.
          This operation is the same as RVEA-iGNG. See \cite{RVEA-iGNG} for the detailed explanation.
\end{enumerate}

\subsection{Mating Selection Strategy based on clustering}\label{sec:ClusterBasedReproduction}


CA generates networks by using the solutions obtained during the search process. If PFs are disconnected or unexplored regions on the PF remain, CA generates several disconnected networks. RVEA-CA uses the information of networks in CA (i.e., the position of nodes and their adjacency relationships by edges) for environmental selection and mating selection. In RVEA-CA, reference vectors are defined by the position of nodes in the networks. Therefore, the reference vectors are efficiently generated near the solutions. Moreover, relationships between reference vectors can be defined by using the adjacency relationships of networks. Based on the relationships between reference vectors, RVEA-CA selects a pair of solutions that will benefit the search process. As a result, the mating selection of RVEA-CA achieves a more efficient local search than that of a random selection-based approach.

\begin{figure}[t]
    \centering
    \begin{minipage}[b]{0.8\linewidth}
        \centering
        \includegraphics[width=\linewidth]{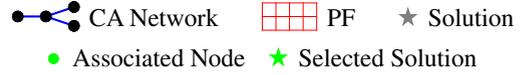}
    \end{minipage}\\
    \begin{minipage}[b]{.6\linewidth}
        \centering
        \includegraphics[width=\linewidth]{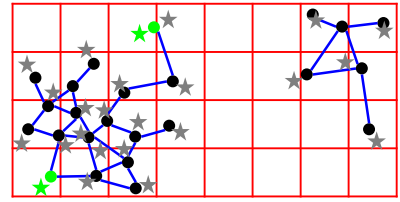}
        \subcaption{Intra-cluster mating}\label{fig:CX}
    \end{minipage}\\
    \vspace{7pt}
    \begin{minipage}[b]{.6\linewidth}
        \centering
        \includegraphics[width=\linewidth]{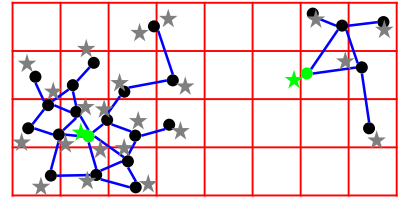}
        \subcaption{Inter-cluster mating}\label{fig:NX}
    \end{minipage}
    \caption{Example of two mating selection strategies based on networks}\label{fig:CXNX}
\end{figure}

Here, two mating selection strategies (i.e., intra-cluster mating and inter-cluster mating) can be considered depending on the organization of networks (i.e., clusters) as follows: i) the first solution is randomly selected from the entire population, and then the second solution is randomly selected from a cluster to which the first solution belongs, and ii) the first solution is randomly selected from the entire population, and then from the other clusters, the nearest neighbor solution of the first solution is selected as the second solution. RVEA-CA randomly uses these two operations on 50\% for each. Fig. \ref{fig:CXNX} illustrates the above operations. Fig. \ref{fig:CX} shows the intra-cluster mating strategy that aims for convergence by searching local regions intensively. Fig. \ref{fig:NX} shows the inter-cluster mating strategy that aims for diversity by searching unexplored regions.
Note that each solution is associated with the node (i.e., a reference vector) with the smallest angle.





Algorithm \ref{alg:ClusterBasedReproduction} summarizes the reproduction process in RVEA-CA.

\begin{algorithm}[htpb]
    \DontPrintSemicolon
    \caption{\newline Cluster Based Reproduction}
    \label{alg:ClusterBasedReproduction}
    \KwIn{\\
        \begin{itemize}
            \setlength{\itemsep}{-0.6mm}
            \item[-] objective function: $ \boldsymbol{f} $,
            \item[-] population: $P = \left\{\boldsymbol{p}_{1},\boldsymbol{p}_{2},\dots,\boldsymbol{p}_{N}\right\}$,
            \item[-] topological network: $G = (\boldsymbol{Y}, E)$,
            \item[-] and estimation of ideal point: $\boldsymbol{z}^{\min}$.
        \end{itemize}
    }
    \KwOut{
        \begin{itemize}
            \setlength{\itemsep}{-0.6mm}
            \item[-] offsprings: $ O $.
        \end{itemize}
    }
    \SetKwBlock{Begin}{function}{end function}
    \Begin(\text{$\operatorname*{ClusterBasedReproduction}_{\boldsymbol{f}}(P,G,\boldsymbol{z}^{\min})$}){
    \uIf{$\boldsymbol{Y}=\emptyset$}{
    $O \gets \operatorname*{Reproduction}_{\boldsymbol{f}}(P, \boldsymbol{z}^{\min})$\\
    }\Else{
    $\boldsymbol{L} = \left(l_{1},l_{2},\dots,l_{N}\right) \gets \operatorname*{Predict}_{\boldsymbol{f}}(P, G, \boldsymbol{z}^{\min})$\\
    $O \gets \emptyset$\\
    \For{$r \gets \operatorname{RandStream}([1,N],N)$}{
        \uIf{$\left|\left\{n \in \left\{1,\dots,N\right\} \mid l_{n} = l_{r}\right\}\right| > 1$}{
            $s \gets \operatorname{Rand}()$\\
            \uIf{$s < \frac{1}{2} \land |\operatorname{Unique}(\boldsymbol{L})| > 0$}{
                $Q \gets \{\boldsymbol{p}_{n} \in P \mid l_{n} \neq l_{r}\}$\\
                $\boldsymbol{q} \gets \argmin_{\boldsymbol{q} \in Q} ||\boldsymbol{f}(\boldsymbol{q}) - \boldsymbol{f}(\boldsymbol{p}_{r})||$
            }\Else{
                $Q \gets \{\boldsymbol{p}_{n} \in P \mid l_{n} = l_{r}\}$\\
                $Q \gets Q \setminus \{\boldsymbol{p}_{r}\}$\\
                $\boldsymbol{q} \gets \operatorname{RandomSample}(Q)$\\
            }
            $\boldsymbol{o} \gets \operatorname{Crossover}(\boldsymbol{p}_{r},\boldsymbol{q})$\\
            $\boldsymbol{o}^{\prime} \gets \operatorname{Mutate}(\boldsymbol{o})$\\
            $O \gets O \cup \left\{\boldsymbol{o}^{\prime}\right\}$\\
        }\Else{
            $\boldsymbol{o} \gets \operatorname{Mutate}(\boldsymbol{p}_{r})$\\
            $O \gets O \cup \left\{\boldsymbol{o}\right\}$\\
        }
    }
    }
    \Return{ \upshape $O$}.
    }
\end{algorithm}

\subsection{Learning Process of CA and Automatic Parameter Adjusting}

The collection of nodes in the network from CA is utilized as the reference vector set of RVEA-CA, and is used in environment selection to select the most promising solution in the local region around each reference vector.
Here, when a new non-dominated solution is generated, its novelty is determined by the vigilance test of CA.
If it has novelty, it is inserted as a new node in the network. If not, the first and second winner nodes are updated according to the similarity, respectively.
By this combination of the searching framework of decomposition-based MOEAs and the learning process of CA, the proposed algorithm quickly and adaptively increases the selection pressure on unexplored regions, thus achieving efficient search.

The characteristics of the network output by CA depend on the distribution of the training data and the hyperparameters $\lambda$ and $V$.
In particular, the similarity threshold $V$ strongly affects the number of nodes, and there is no robust setting in the general or practical range.
Usually, a grid search is required for such sensitive and data-dependent parameters.
When the reference vectors are insufficient, i.e., the similarity threshold $V$ is too large, the subspace allocated to each sub-problem also becomes larger, which may decrease the efficiency of the search.
On the other hand, when the reference vectors are excessive, i.e., the similarity threshold $V$ is too small, the distribution of nodes is strongly influenced by the local density of the training data, which will bias the search and reduce the efficiency of the search as well.
In addition, the computational costs for environment selection in decomposition-based MOEAs are proportional to the number of reference vectors.

The key to solving the above issue is the property that the number of nodes generated by CA is monotonically decreasing concerning the similarity threshold $V$.
Although it is data-dependent how many nodes CA generates for a given $V$, this monotonicity holds for any training data.
Therefore, the proposed algorithm introduces an automatic adjustment strategy for the similarity threshold $V$ using a pseudo-binary search method.
This consists of the following three steps:

\begin{enumerate}
    \item
          Initialize the current similarity threshold $V$ with the value inherited from the previous generation, and initialize the upper bound $V_{u}$ of $V$ with 1 and the lower bound $V_{l}$ of $V$ with 0.
    \item
          Obtain a collection of nodes from the CA network by learning the training data with the current similarity threshold $V$.
    \item
          If the number of nodes is within the acceptable range, which in this paper is $[0.75N,1.25N]$ for the population size $N$, output a collection of nodes and the current similarity threshold $V$ and complete this procedure.
          If the number of nodes is larger than the acceptable range, set lower bound $V_{l}$ to the current similarity threshold $V$. If the number of nodes is smaller than the acceptable range, set upper bound $V_{u}$ to the current similarity threshold $V$.
          Then update the current similarity threshold $V$ as follows, and return to step 2.
          \begin{equation}
              V \gets \frac{V_{u}+V_{l}}{2}.
          \end{equation}
\end{enumerate}

In this paper, we limit the number of iterations to five to balance the computational cost and the accuracy of $V$.
After the midpoint of the search process, the number of inputs increases, the distribution of training data becomes stable, and then the correspondence between the number of nodes and $V$ also becomes stable.
Therefore, this adjusting process is complete with no iteration, and we can estimate that the additional computational cost is small compared to the case of fixing $V$.

While RVEA-iGNG inherits the network that approximates the distribution of the training data from the previous generation, RVEA-CA does not inherit the network and learns it from a state with no nodes.
In this way, RVEA-CA fundamentally solves the problem of which nodes should be deleted, which is one of the challenges in applying clustering algorithms to decomposition-based MOEAs.
Clustering algorithms either do not have a node deletion mechanism or, if they do, their purpose is often to remove noise.
Therefore, they do not meet the requirement of removing nodes that are not necessary for approximating PF shapes generated in past generations.
RVEA-iGNG attempts to alleviate this problem by introducing an additional node deletion mechanism to GNG~\cite{RVEA-iGNG}.
On the other hand, CA can quickly converge to the distribution of the solutions, which is the training data, even when learning from a state where no nodes exist.
For this reason, RVEA-CA employs CA and learns from a state where there are no nodes in each generation, making the node deletion mechanism itself unnecessary.

\begin{algorithm}[tb]
    \DontPrintSemicolon
    \caption{\newline Learning and Adjusting Procedure of CA}
    \label{alg:AdaptiveTrainCA}
    \KwIn{\\
        \begin{itemize}
            \setlength{\itemsep}{-0.6mm}
            \item[-] training instances: $ \boldsymbol{X} $,
            \item[-] an interval for adapting $ \sigma $: $ \lambda $,
            \item[-] a similarity threshold: $ V $,
            \item[-] and population size: $N$.
        \end{itemize}
    }
    \KwOut{
        \begin{itemize}
            \setlength{\itemsep}{-0.6mm}
            \item[-] node collection: $ \boldsymbol{Y} $,
            \item[-] edge set: $ E $,
            \item[-] and adapted similarity threshold: $ V $.
        \end{itemize}
    }
    \SetKwBlock{Begin}{function}{end function}
    \Begin(\text{AdaptiveTrainCA($\boldsymbol{X}$, $\lambda$, $V$, $N$)}){
        \uIf{$\operatorname{size}\left(\boldsymbol{X}\right) \leq 1.25N$}{
            $V \gets 0.1$ \\
            $\boldsymbol{Y}, E \gets \operatorname{TrainCA}\left(\boldsymbol{X},\left(\emptyset,\emptyset\right),\emptyset,\emptyset,\lambda,V\right)$ \\ 
        }\Else{
            $V_{u} \gets 1$ \\
            $V_{l} \gets 0$ \\
            \For{$i \gets 1$ \KwTo $5$}{
                $\boldsymbol{Y}, E \gets \operatorname{TrainCA}\left(\boldsymbol{X},\left(\emptyset,\emptyset\right),\emptyset,\emptyset,\lambda,V\right)$ \\
                \uIf{$0.75N \leq \operatorname{size}(\boldsymbol{Y}) \leq 1.25N$}{
                    break \\
                }\uElseIf{$\operatorname{size}(\boldsymbol{Y}) < 0.75N$}{
                    $V_u \gets V$ \\
                }\Else{
                    $V_l \gets V$ \\
                }
                $V \gets \frac{V_u + V_l}{2}$ \\
            }
        }
        \Return{ \upshape $\boldsymbol{Y}$, $E$, and $V$}.
    }
\end{algorithm}

\section{Computational Experiments}\label{sec:Experiments}

In this section, we compare search performance for MaOPs of RVEA-CA, RVEA~\cite{RVEA}, RVEA-iGNG~\cite{RVEA-iGNG}, DEA-GNG~\cite{DEA-GNG}, A-NSGA-III~\cite{NSGA-III-1}, VaEA~\cite{VaEA}, AR-MOEA~\cite{AR-MOEA}, MOEA/D-AWA~\cite{MOEAD-AWA}, and AdaW~\cite{AdaW} by computational experiments.
All algorithms and problems are implemented on the PlatEMO platform~\cite{PlatEmo}.

\subsection{Test problems}

The search performance of RVEA-CA is examined by using MaF test suite~\cite{MaF}.
Since MaF14 and MaF15 are large-scale problems, we exclude them from experiments.
The MaF test suite consists of various types of problems selected from several other problem suites to evaluate the overall search performance of algorithms for MaOPs. 
The properties of problems in MaF test suite are listed in Table \ref{tbl:MaF}.

\begin{table}[t]
    \centering
    \caption{MaF test suite~\cite{MaF}}
    \label{tbl:MaF}
    \small
    \begin{tabular*}{\linewidth}{llp{0.55\linewidth}}\toprule
        Problem & Base       & Properties                                  \\\midrule
        MaF1    & IDTLZ1     & Linear                                      \\
        MaF2    & DTLZ2BZ    & Concave                                     \\
        MaF3    & CDTLZ3     & Convex, multimodal                          \\
        MaF4    & IDTLZ3     & Concave, multimodal                         \\
        MaF5    & CDTLZ4     & Convex, biased                              \\
        MaF6    & DTLZ5(I,M) & Concave, degenerate                         \\
        MaF7    & DTLZ7      & Mixed, disconnected, Multimodal             \\
        MaF8    & MPDMP      & Linear, degenerate                          \\
        MaF9    & MLDMP      & Linear, degenerate                          \\
        MaF10   & WFG1       & Mixed, biased                               \\
        MaF11   & WFG2       & Convex, disconnected, nonseparable          \\
        MaF12   & WFG9       & Concave, nonseparable,\newline biased deceptive     \\
        MaF13   & PF7        & Concave, unimodal, nonseparable, degenerate \\\bottomrule
    \end{tabular*}
\end{table}

\begin{table*}[htb]
    \centering
    \caption{Population size and number of evaluations for each number of objectives}
    \label{tbl:settings}
    \begin{tabular}{ccccccc}
        \toprule
        Number of Objectives  & 5      & 8      & 10      & 13      & 15      & 20      \\ \midrule
        Population Size       & 210    & 240    & 230     & 182     & 240     & 230     \\
        Number of Evaluations & 50,190 & 80,160 & 100,050 & 130,130 & 150,000 & 200,100 \\ \toprule
    \end{tabular}
\end{table*}

\subsection{Performance Indicators}

In this paper, we adopt hypervolume (HV)~\cite{HV} and inverted generational distance plus (IGD${}^+$)~\cite{IGDp} to evaluate the search performances of MOEAs by applying them to the final population.
HV and IGD${}^+$ are widely used performance indicators that evaluate the convergence and diversity of a given population simultaneously.
Note that we normalize objective function values of populations before calculating performance indicators.

HV is the size of the area that is dominated by a given population and bounded by a reference point. The HV calculation for the given population $P$ is defined as follows:
\begin{equation}\label{eq:HV}
    \operatorname{HV}(P) = \mathcal{L}\left(\bigcup_{\boldsymbol{p}\in P}\left(\prod_{m=1}^{M} \left(f_{m}\left(\boldsymbol{p}\right), q_{m}\right]\right)\right),
\end{equation}
where $\mathcal{L}(\cdot)$ is Lebesgue measure, and $M$-dimensional vector $\boldsymbol{q} = \left(q_{1},q_{2},\dots, q_{M}\right)$ is the reference point of HV.
As you can see from (\ref{eq:HV}), only one reference point is needed to calculate HV, and considering characteristics of HV for MaOPs, we set $\boldsymbol{q} = (1.5,1.5,\dots,1.5)$.
To reduce computational time, we use the Monte Carlo method and parallel computing on a GPU to estimate the HV values.
This code works the same as the version implemented in PlatEMO.
The larger the HV value for a population, the better the population is regarded.

IGD${}^+$ is the average of the distance from each reference point to the area dominated by the given population, and the IGD${}^+$ value for the given population $P$ is defined as follows:
\begin{equation}\label{eq:IGDPlus}
    \operatorname{IGD}^{+}\left(P\right) = \frac{1}{|R|}\sum_{\boldsymbol{r}\in R}\min_{\boldsymbol{p} \in P}\left\{d^{+}\left(\boldsymbol{r}, \boldsymbol{f}\left(\boldsymbol{p}\right)\right)\right\},
\end{equation}
where $R$ is the reference point set of IGD${}^+$, and $d^{+}$ is the modified Euclidean distance defined as follows:
\begin{equation}
    d^{+}\left(\boldsymbol{r}, \boldsymbol{f}\left(\boldsymbol{p}\right)\right) = \sqrt{\sum_{m=1}^{M}\left(\max\left\{f_{m}\left(\boldsymbol{p}\right) - r_{m}, 0\right\}\right)^2}.
\end{equation}
The reference point set of IGD${}^+$ is typically uniformly distributed on the true PF.
In this paper, we use a normalized version of the reference point sets provided by PlatEMO.
The smaller the IGD${}^+$ value for a population, the better the population is regarded.

\subsection{Parameter Settings}

In this paper, we use 5-, 8-, 10-, 15-, and 20-objective test problems to compare the search performances of algorithms.
We specify the population size and the number of evaluations in Table \ref{tbl:settings}.
In addition, we set the number of runs to 31.

For RVEA-CA, $\lambda$ is set to 100.
For RVEA, $\alpha$ is set to 2, and $f_{r}$ is set to 0.1.
For RVEA-iGNG, we set parameters as $\varepsilon_{b} = 0.2$, $\varepsilon_{n} = 0.006$, $\alpha = 0.5$, $\alpha_{\max} = 50$, $d = 0.995$, $\lambda = 50$, $f_{r} = 0.1$.
For DEA-GNG, $\epsilon$ is set to $0.15\pi$, the size of the archive set is set to $MN$, where $M$ is the number of objectives and $N$ is the population size.
For MOEA/D-AWA, $rate\_update\_weight$ is set to 0.05, $rate\_evol$ is set to 0.8, and $wag$ is set to 100.

In this paper, we use Simulated Binary Crossover (SBX)~\cite{SBX} as the crossover operator and Polynomial Mutation (PM)~\cite{PM} as the mutation operator.
For SBX, the SBX distribution index $\eta_{c}$ is set to 20, and the crossover probability $p_{c}$ is set to 1.
For PM, the PM distribution index $\eta_{m}$ is set to 20, and the mutation probability $p_{m}$ is set to $\frac{1}{D}$, where $D$ is number of dimensions of the decision space.

\subsection{Experimental Results}

The experimental results are showed in Tables \ref{tbl:HV1}-\ref{tbl:HV2} for HV and Tables \ref{tbl:IGDP1}-\ref{tbl:IGDP2} for IGD${}^+$.
Each column in Tables \ref{tbl:HV1}-\ref{tbl:IGDP2} represents each algorithm, and each row represents each test instance.
Each cell in Tables \ref{tbl:HV1}-\ref{tbl:IGDP2} shows the median and standard deviation of the values of HV and IGD${}^+$ in 31 runs.
In each row, the algorithm with the best median value is highlighted.
In addition, the result of comparison with RVEA-CA at a 0.05 level by Wilcoxon's rank-sum test~\cite{Wilcoxon} is shown on the right end of each cell in each row except for RVEA-CA.
``$+$'' and ``$-$'' mean that RVEA-CA performs significantly better than and significantly worse than each algorithm, respectively, and ``$\approx$'' implies a non-significant difference in performance between RVEA-CA and its competitor.

Moreover, Table \ref{tbl:summaryOfResult} aggregates the number of the significant difference in median value shown in Tables \ref{tbl:HV1}-\ref{tbl:IGDP2}.
``$+$'', ``$\approx$'', and ``$-$'' have the same meaning in Tables \ref{tbl:HV1}-\ref{tbl:IGDP2}.
In addition, Table \ref{tbl:summaryOfResult} also aggregates the result of comparisons for each property described in Table \ref{tbl:MaF}, i.e., Linear, Concave, Convex, Mixed, Degenerate, Disconnected, Biased, Multimodal, and Nonseparable.
If RVEA-CA outperforms its competitor in terms of both HV and IGD${}^+$, the corresponding cell in Table \ref{tbl:summaryOfResult} is highlighted, and if RVEA-CA outperforms its competitor in terms of either HV or IGD${}^+$, the result in the cell in Table \ref{tbl:summaryOfResult} is emphasized in bold.
Note that because every test problem has multiple properties, there is some overlap between aggregations.

\begin{sidewaystable*}[htbp]
    \centering
    \caption{Median value and standard deviation of 31 runs on HV (I)}
    \scalebox{0.66}{
        \begin{tabular}{ccccccccccc}
            \toprule
            Problem               & $M$ & RVEA-CA                  & RVEA                          & RVEA-iGNG                          & DEA-GNG                       & A-NSGA-III                    & VaEA                               & AR-MOEA                            & MOEA/D-AWA                         & AdaW                          \\
            \midrule
            \multirow{6}{*}{MaF1} & 5   & 1.5322e+0 (1.25e-2)      & 8.9898e-1 (3.65e-2) $+$       & 1.5409e+0 (7.22e-3) $-$            & 1.4257e+0 (4.08e-2) $+$       & 9.8867e-1 (8.49e-2) $+$       & 1.5114e+0 (7.69e-3) $+$            & 1.4976e+0 (1.19e-2) $+$            & 1.2926e+0 (8.12e-3) $+$            & \hl{1.5637e+0 (8.63e-3) $-$}  \\
                                  & 8   & 5.5484e-1 (3.21e-2)      & 1.4914e-1 (2.95e-2) $+$       & 5.3281e-1 (1.76e-2) $+$            & 5.4188e-1 (2.64e-2) $\approx$ & 2.6370e-1 (2.11e-2) $+$       & 6.6295e-1 (7.38e-3) $-$            & 6.0538e-1 (7.86e-3) $-$            & 2.5941e-1 (2.80e-2) $+$            & \hl{6.9650e-1 (6.16e-3) $-$}  \\
                                  & 10  & 2.2564e-1 (2.95e-2)      & 4.4695e-2 (9.97e-3) $+$       & 2.1705e-1 (3.38e-2) $\approx$      & 2.4970e-1 (1.40e-2) $-$       & 1.1073e-1 (1.06e-2) $+$       & 3.3753e-1 (4.03e-3) $-$            & 3.0688e-1 (6.22e-3) $-$            & 9.9665e-2 (1.33e-2) $+$            & \hl{3.4277e-1 (6.81e-3) $-$}  \\
                                  & 13  & 3.8302e-2 (4.23e-3)      & 6.6148e-3 (1.42e-3) $+$       & 3.1556e-2 (7.95e-3) $+$            & 5.0567e-2 (6.72e-3) $-$       & 1.9561e-2 (1.79e-3) $+$       & 8.0269e-2 (1.79e-3) $-$            & 7.4947e-2 (3.34e-3) $-$            & 1.9039e-2 (4.35e-3) $+$            & \hl{8.1007e-2 (2.29e-3) $-$}  \\
                                  & 15  & 1.6342e-2 (2.64e-3)      & 2.5394e-3 (5.14e-4) $+$       & 1.0609e-2 (2.22e-3) $+$            & 2.1879e-2 (2.67e-3) $-$       & 8.5952e-3 (9.28e-4) $+$       & \hl{4.1321e-2 (1.92e-3) $-$}       & 3.7880e-2 (2.17e-3) $-$            & 8.4416e-3 (2.19e-3) $+$            & 4.0854e-2 (2.24e-3) $-$       \\
                                  & 20  & 9.4503e-4 (2.69e-4)      & 1.4413e-4 (2.26e-5) $+$       & 7.3666e-4 (2.03e-4) $+$            & 1.3418e-3 (3.43e-4) $-$       & 5.0930e-4 (7.28e-5) $+$       & \hl{3.2871e-3 (4.05e-5) $-$}       & 2.8259e-3 (1.14e-3) $-$            & 4.4019e-4 (8.47e-5) $+$            & 2.7086e-3 (7.08e-4) $-$       \\
            \hline
            \multirow{6}{*}{MaF2} & 5   & 3.7377e+0 (8.56e-3)      & 3.4840e+0 (3.70e-2) $+$       & \hl{3.7456e+0 (8.44e-3) $-$}       & 3.4991e+0 (4.91e-2) $+$       & 2.7954e+0 (2.50e-1) $+$       & 3.6138e+0 (3.65e-2) $+$            & 3.5740e+0 (2.35e-2) $+$            & 3.5753e+0 (1.37e-2) $+$            & 3.6358e+0 (4.18e-2) $+$       \\
                                  & 8   & 1.2915e+1 (7.77e-2)      & 9.8941e+0 (5.79e-1) $+$       & \hl{1.3013e+1 (6.02e-2) $-$}       & 1.1657e+1 (2.65e-1) $+$       & 1.1142e+1 (9.41e-1) $+$       & 1.2678e+1 (1.02e-1) $+$            & 1.1547e+1 (2.05e-1) $+$            & 1.1975e+1 (1.00e-1) $+$            & 1.2086e+1 (2.15e-1) $+$       \\
                                  & 10  & 2.8775e+1 (1.71e-1)      & 2.4282e+1 (1.42e+0) $+$       & \hl{2.8904e+1 (1.89e-1) $-$}       & 2.5815e+1 (5.68e-1) $+$       & 2.4983e+1 (1.88e+0) $+$       & 2.8151e+1 (1.90e-1) $+$            & 2.7131e+1 (3.39e-1) $+$            & 2.6152e+1 (2.42e-1) $+$            & 2.6409e+1 (4.57e-1) $+$       \\
                                  & 13  & \hl{9.5553e+1 (6.63e-1)} & 5.1698e+1 (2.57e+0) $+$       & 9.5247e+1 (6.71e-1) $+$            & 8.4548e+1 (2.47e+0) $+$       & 6.7297e+1 (1.14e+1) $+$       & 9.3150e+1 (7.65e-1) $+$            & 7.4395e+1 (3.07e+0) $+$            & 7.9367e+1 (3.40e+0) $+$            & 7.9025e+1 (3.48e+0) $+$       \\
                                  & 15  & 2.1537e+2 (1.91e+0)      & 1.1494e+2 (9.70e+0) $+$       & \hl{2.1543e+2 (1.60e+0) $\approx$} & 1.8954e+2 (5.51e+0) $+$       & 1.6990e+2 (1.67e+1) $+$       & 2.0977e+2 (1.60e+0) $+$            & 1.7430e+2 (6.79e+0) $+$            & 1.7770e+2 (7.83e+0) $+$            & 1.7264e+2 (9.41e+0) $+$       \\
                                  & 20  & 1.6210e+3 (1.54e+1)      & 1.0100e+3 (1.29e+2) $+$       & \hl{1.6322e+3 (1.31e+1) $-$}       & 1.4451e+3 (4.97e+1) $+$       & 1.2671e+3 (1.34e+2) $+$       & 1.5772e+3 (1.38e+1) $+$            & 1.5192e+3 (3.03e+1) $+$            & 1.3445e+3 (5.60e+1) $+$            & 1.1994e+3 (7.54e+1) $+$       \\
            \hline
            \multirow{6}{*}{MaF3} & 5   & 3.6890e+0 (3.58e+0)      & 0.0000e+0 (0.00e+0) $+$       & 6.8040e+0 (2.21e+0) $-$            & 3.0923e+0 (3.71e+0) $\approx$ & 3.3474e+0 (3.30e+0) $\approx$ & 4.1056e+0 (3.37e+0) $\approx$      & 6.2937e+0 (2.57e+0) $-$            & \hl{7.5647e+0 (3.05e-2) $-$}       & 7.3310e+0 (1.36e+0) $-$       \\
                                  & 8   & 9.4898e+0 (1.05e+1)      & 1.2900e+1 (1.27e+1) $\approx$ & 2.0987e+1 (8.97e+0) $-$            & 5.3811e+0 (9.57e+0) $+$       & 0.0000e+0 (0.00e+0) $+$       & 0.0000e+0 (0.00e+0) $+$            & 2.4793e+1 (4.60e+0) $-$            & \hl{2.5625e+1 (2.27e-3) $-$}       & 2.5574e+1 (8.99e-2) $-$       \\
                                  & 10  & 3.6262e+1 (2.60e+1)      & 4.0313e+1 (2.61e+1) $\approx$ & 3.5474e+1 (2.58e+1) $\approx$      & 1.5786e+1 (2.53e+1) $+$       & 0.0000e+0 (0.00e+0) $+$       & 0.0000e+0 (0.00e+0) $+$            & 5.5794e+1 (1.04e+1) $\approx$      & \hl{5.7664e+1 (1.28e-3) $\approx$} & 5.7538e+1 (2.49e-1) $\approx$ \\
                                  & 13  & \hl{1.9461e+2 (1.26e-2)} & 1.9452e+2 (1.41e-1) $+$       & 1.9443e+2 (7.59e-1) $+$            & 7.9141e+1 (9.05e+1) $+$       & 0.0000e+0 (0.00e+0) $+$       & 0.0000e+0 (0.00e+0) $+$            & 1.6553e+2 (6.82e+1) $+$            & 1.9457e+2 (2.43e-2) $+$            & 1.9443e+2 (3.78e-1) $+$       \\
                                  & 15  & \hl{4.3789e+2 (1.79e-2)} & 4.3528e+2 (1.42e+1) $+$       & 4.3777e+2 (3.78e-1) $+$            & 2.6899e+2 (2.04e+2) $+$       & 0.0000e+0 (0.00e+0) $+$       & 0.0000e+0 (0.00e+0) $+$            & 2.8993e+2 (2.06e+2) $+$            & 4.3780e+2 (7.81e-2) $+$            & 4.3701e+2 (1.15e+0) $+$       \\
                                  & 20  & 3.3252e+3 (9.20e-2)      & 3.3241e+3 (1.24e+0) $+$       & 3.3250e+3 (2.64e-1) $+$            & 2.9842e+3 (8.08e+2) $+$       & 0.0000e+0 (0.00e+0) $+$       & 0.0000e+0 (0.00e+0) $+$            & 2.7269e+3 (1.26e+3) $+$            & \hl{3.3252e+3 (5.30e-2) $-$}       & 3.3189e+3 (6.56e+0) $+$       \\
            \hline
            \multirow{6}{*}{MaF4} & 5   & 2.1801e+0 (1.02e+0)      & 1.4291e+0 (5.55e-1) $+$       & 2.7237e+0 (6.26e-1) $-$            & 2.5965e+0 (5.01e-1) $\approx$ & 1.2314e+0 (1.18e+0) $+$       & 2.4999e+0 (4.52e-1) $\approx$      & 2.5989e+0 (4.90e-1) $\approx$      & 1.7340e+0 (6.11e-2) $+$            & \hl{2.8781e+0 (1.42e-1) $-$}  \\
                                  & 8   & 2.0141e+0 (2.59e-1)      & 2.4616e-1 (7.74e-2) $+$       & 1.8657e+0 (3.29e-1) $+$            & 1.8470e+0 (1.87e-1) $+$       & 9.0039e-1 (5.93e-1) $+$       & 1.8710e+0 (2.03e-1) $+$            & 8.7577e-1 (8.19e-2) $+$            & 5.9364e-1 (1.77e-1) $+$            & \hl{2.1535e+0 (1.73e-1) $-$}  \\
                                  & 10  & \hl{1.3634e+0 (1.51e-1)} & 9.8265e-2 (3.01e-2) $+$       & 1.1120e+0 (2.50e-1) $+$            & 1.2698e+0 (1.64e-1) $+$       & 7.0424e-1 (3.28e-1) $+$       & 1.2961e+0 (1.05e-1) $+$            & 3.1103e-1 (3.79e-2) $+$            & 3.1952e-1 (1.21e-1) $+$            & 1.3596e+0 (1.62e-1) $\approx$ \\
                                  & 13  & \hl{5.8272e-1 (4.49e-2)} & 1.5844e-2 (9.35e-3) $+$       & 4.7315e-1 (5.44e-2) $+$            & 4.2858e-1 (5.08e-2) $+$       & 2.5195e-1 (8.49e-2) $+$       & 5.0453e-1 (3.37e-2) $+$            & 5.3490e-2 (5.04e-3) $+$            & 1.4595e-1 (7.24e-2) $+$            & 5.0688e-1 (6.60e-2) $+$       \\
                                  & 15  & 2.7108e-1 (8.90e-2)      & 6.1862e-3 (3.49e-3) $+$       & 2.0925e-1 (5.99e-2) $+$            & 2.7135e-1 (3.88e-2) $\approx$ & 1.5741e-1 (5.86e-2) $+$       & \hl{2.8880e-1 (3.76e-2) $\approx$} & 2.0949e-2 (2.80e-3) $+$            & 9.8873e-2 (3.48e-2) $+$            & 2.6816e-1 (4.37e-2) $\approx$ \\
                                  & 20  & 3.1876e-2 (1.90e-2)      & 3.9911e-4 (1.53e-4) $+$       & 1.4901e-2 (7.66e-3) $+$            & 4.3252e-2 (7.17e-3) $-$       & 1.6808e-2 (9.85e-3) $+$       & \hl{4.8165e-2 (5.64e-3) $-$}       & 1.4770e-3 (1.85e-4) $+$            & 3.9088e-3 (2.79e-3) $+$            & 3.8692e-2 (5.73e-3) $\approx$ \\
            \hline
            \multirow{6}{*}{MaF5} & 5   & 7.3896e+0 (4.16e-2)      & 7.3784e+0 (1.51e-1) $+$       & 7.3803e+0 (5.58e-2) $+$            & 7.3017e+0 (2.94e-2) $+$       & 7.3031e+0 (7.10e-2) $+$       & 7.3825e+0 (3.52e-3) $+$            & \hl{7.4058e+0 (3.58e-4) $-$}       & 7.0771e+0 (5.08e-1) $+$            & 7.3763e+0 (6.80e-2) $\approx$ \\
                                  & 8   & \hl{2.5550e+1 (4.14e-3)} & 2.5515e+1 (4.40e-2) $+$       & 2.5528e+1 (4.55e-2) $+$            & 2.5462e+1 (3.56e-2) $+$       & 2.4408e+1 (5.87e-1) $+$       & 2.5525e+1 (6.40e-3) $+$            & 2.5547e+1 (1.46e-3) $+$            & 2.5235e+1 (3.33e-1) $+$            & 2.5537e+1 (1.16e-2) $+$       \\
                                  & 10  & 5.7614e+1 (4.28e-3)      & 5.7603e+1 (1.92e-2) $+$       & 5.7605e+1 (1.34e-2) $+$            & 5.7498e+1 (5.19e-2) $+$       & 5.1404e+1 (3.00e+0) $+$       & 5.7578e+1 (1.81e-2) $+$            & \hl{5.7629e+1 (1.57e-3) $-$}       & 5.7279e+1 (3.71e-1) $+$            & 5.7610e+1 (8.04e-3) $+$       \\
                                  & 13  & 1.9458e+2 (7.41e-3)      & 1.9434e+2 (4.28e-1) $+$       & 1.9450e+2 (5.75e-2) $+$            & 1.9457e+2 (2.05e-2) $\approx$ & 1.4755e+2 (1.57e+1) $+$       & 1.9458e+2 (4.41e-3) $+$            & \hl{1.9459e+2 (4.74e-3) $-$}       & 1.9336e+2 (2.53e+0) $+$            & 1.9456e+2 (1.64e-2) $+$       \\
                                  & 15  & 4.3786e+2 (8.76e-3)      & 4.3781e+2 (3.00e-2) $+$       & 4.3779e+2 (4.76e-2) $+$            & \hl{4.3788e+2 (4.73e-3) $-$}  & 3.5418e+2 (4.92e+1) $+$       & 4.3784e+2 (2.75e-2) $+$            & 4.3787e+2 (4.83e-3) $-$            & 4.3668e+2 (3.01e+0) $+$            & 4.3786e+2 (1.09e-2) $\approx$ \\
                                  & 20  & 3.3252e+3 (2.76e-2)      & 3.3251e+3 (1.44e-1) $+$       & 3.3250e+3 (1.19e-1) $+$            & \hl{3.3253e+3 (3.86e-3) $-$}  & 2.1983e+3 (6.14e+2) $+$       & 3.3251e+3 (1.92e-1) $\approx$      & 3.3253e+3 (2.01e-3) $-$            & 3.3238e+3 (2.56e+0) $+$            & 3.3252e+3 (1.08e-2) $-$       \\
            \hline
            \multirow{6}{*}{MaF6} & 5   & 3.5053e+0 (1.02e-3)      & 3.3423e+0 (3.18e-2) $+$       & 3.5054e+0 (5.77e-4) $\approx$      & \hl{3.5060e+0 (1.97e-4) $-$}  & 3.4822e+0 (1.50e-2) $+$       & 3.5053e+0 (1.80e-4) $-$            & 3.5053e+0 (1.85e-4) $-$            & 3.3767e+0 (2.28e-2) $+$            & 3.4990e+0 (1.72e-2) $+$       \\
                                  & 8   & 7.9351e+0 (4.76e+0)      & 9.8228e+0 (1.81e+0) $-$       & 1.0747e+1 (9.33e-3) $\approx$      & 7.3072e+0 (5.08e+0) $+$       & 7.5967e+0 (4.94e+0) $+$       & 7.9776e+0 (4.78e+0) $-$            & \hl{1.0750e+1 (4.23e-4) $\approx$} & 1.0690e+1 (7.19e-3) $-$            & 1.0736e+1 (5.37e-2) $-$       \\
                                  & 10  & 3.0414e+0 (8.03e+0)      & 2.3065e+1 (1.12e-1) $-$       & \hl{2.3597e+1 (1.88e-2) $-$}       & 7.6141e-1 (4.24e+0) $\approx$ & 0.0000e+0 (0.00e+0) $+$       & 0.0000e+0 (0.00e+0) $+$            & 1.7233e+1 (1.04e+1) $-$            & 2.3458e+1 (5.06e-2) $-$            & 2.3565e+1 (1.11e-1) $-$       \\
                                  & 13  & 6.5443e+0 (2.09e+1)      & 7.7021e+1 (2.65e-1) $-$       & 7.5782e+1 (1.41e+1) $-$            & 3.8496e+0 (1.49e+1) $\approx$ & 8.9505e+0 (2.36e+1) $\approx$ & 1.4508e+1 (2.75e+1) $\approx$      & 7.6817e+1 (1.70e+0) $-$            & \hl{7.8101e+1 (9.28e-2) $-$}       & 7.8092e+1 (5.92e-1) $-$       \\
                                  & 15  & 4.8716e-5 (2.43e-4)      & 1.7286e+2 (5.06e-1) $-$       & 1.5688e+2 (4.98e+1) $-$            & 3.4141e+0 (1.90e+1) $\approx$ & 1.0352e+1 (4.01e+1) $\approx$ & 8.9145e+0 (2.74e+1) $\approx$      & 1.7131e+2 (3.15e+0) $-$            & 1.7464e+2 (1.27e-1) $-$            & \hl{1.7483e+2 (5.62e-1) $-$}  \\
                                  & 20  & 3.0903e+1 (1.34e+2)      & 1.3104e+3 (2.19e-2) $-$       & 4.9980e+2 (6.28e+2) $-$            & 8.0002e+1 (3.10e+2) $\approx$ & 3.8305e+1 (2.13e+2) $-$       & 3.8329e+2 (5.58e+2) $-$            & 1.3088e+3 (9.89e+0) $-$            & \hl{1.3204e+3 (8.65e-1) $-$}       & 1.3203e+3 (4.67e+0) $-$       \\
            \hline
            \multirow{6}{*}{MaF7} & 5   & 3.7163e+0 (1.00e-1)      & 4.2465e+0 (1.75e-2) $-$       & \hl{4.5843e+0 (3.25e-2) $-$}       & 4.3769e+0 (7.41e-2) $-$       & 3.9371e+0 (1.99e-1) $-$       & 4.4423e+0 (1.67e-2) $-$            & 4.5386e+0 (9.69e-3) $-$            & 4.0517e+0 (1.24e-1) $-$            & 4.4981e+0 (6.80e-2) $-$       \\
                                  & 8   & 1.4810e+1 (1.53e-1)      & 1.2813e+1 (1.77e-1) $+$       & \hl{1.4850e+1 (1.24e-1) $\approx$} & 1.4256e+1 (9.49e-2) $+$       & 1.1357e+1 (5.06e-1) $+$       & 1.3656e+1 (8.94e-2) $+$            & 1.3563e+1 (1.24e-1) $+$            & 1.2883e+1 (1.62e-1) $+$            & 1.3701e+1 (1.58e-1) $+$       \\
                                  & 10  & \hl{3.1804e+1 (1.84e-1)} & 2.6096e+1 (7.26e-1) $+$       & 3.1307e+1 (2.25e-1) $+$            & 3.0432e+1 (4.53e-1) $+$       & 2.2710e+1 (2.97e+0) $+$       & 2.9332e+1 (2.36e-1) $+$            & 2.7123e+1 (6.34e-1) $+$            & 2.8951e+1 (1.93e-1) $+$            & 2.9577e+1 (6.08e-1) $+$       \\
                                  & 13  & 9.4136e+1 (2.62e+0)      & 5.8022e+1 (3.02e+1) $+$       & 8.2324e+1 (8.17e+0) $+$            & 8.9107e+1 (2.17e+0) $+$       & 0.0000e+0 (0.00e+0) $+$       & \hl{9.5141e+1 (8.86e-1) $-$}       & 6.6938e+1 (1.29e+1) $+$            & 2.4982e+1 (1.67e+1) $+$            & 8.9230e+1 (5.52e+0) $+$       \\
                                  & 15  & \hl{2.0963e+2 (6.47e+0)} & 6.1157e+1 (4.33e+1) $+$       & 1.3739e+2 (1.68e+1) $+$            & 1.9686e+2 (4.21e+0) $+$       & 0.0000e+0 (0.00e+0) $+$       & 2.0345e+2 (2.31e+0) $+$            & 1.1495e+2 (6.15e+1) $+$            & 3.0393e+1 (2.27e+1) $+$            & 1.8812e+2 (3.57e+1) $+$       \\
                                  & 20  & 1.4611e+3 (2.43e+2)      & 1.3673e+3 (3.26e+1) $+$       & 4.3381e+2 (2.20e+2) $+$            & 1.4059e+3 (6.28e+1) $+$       & 0.0000e+0 (0.00e+0) $+$       & \hl{1.4619e+3 (2.27e+1) $-$}       & 5.6183e+2 (5.31e+2) $+$            & 1.3853e+3 (3.06e+1) $+$            & 4.0280e+1 (2.24e+2) $+$       \\
            \bottomrule
        \end{tabular}
    }
    \\
	\vspace{1mm}
	\footnotesize \raggedright \hspace{4mm}Each column represents each algorithm, and each row represents each test instance.
	
	\hspace{4mm}Each cell shows the median and standard deviation of the evaluations of HV and IGD${}^+$ in 31 runs.
	
	\hspace{4mm}The algorithm with the best median value for each row is highlighted.
	
	\hspace{4mm}The result of comparison with RVEA-CA at a 0.05 level by Wilcoxon's rank-sum test is shown on the right end of each cell in each row except for RVEA-CA.
	
	\hspace{4mm}``$+$'', ``$\approx$'', and ``$-$'' mean that RVEA-CA performs significantly better than, non-significantly better/worse than, and significantly worse than each algorithm, respectively.
    \label{tbl:HV1}
\end{sidewaystable*}

\begin{sidewaystable*}[htbp]
    \centering
    \caption{Median value and standard deviation of 31 runs on HV (II)}
    \scalebox{0.66}{
        \begin{tabular}{ccccccccccc}
            \toprule
            Problem                & $M$ & RVEA-CA                  & RVEA                          & RVEA-iGNG                     & DEA-GNG                       & A-NSGA-III              & VaEA                          & AR-MOEA                       & MOEA/D-AWA                    & AdaW                          \\
            \midrule
            \multirow{6}{*}{MaF8}  & 5   & 2.5994e+0 (1.12e-1)      & 2.2175e+0 (7.29e-2) $+$       & \hl{2.9272e+0 (2.42e-2) $-$}  & 2.8093e+0 (5.76e-2) $-$       & 1.9184e+0 (3.08e-1) $+$ & 2.8970e+0 (3.43e-2) $-$       & 2.8848e+0 (1.59e-2) $-$       & 2.6020e+0 (4.95e-2) $\approx$ & 2.6679e+0 (1.75e-1) $\approx$ \\
                                   & 8   & 4.5647e+0 (1.08e-1)      & 2.3297e+0 (3.22e-1) $+$       & \hl{4.7640e+0 (7.73e-3) $-$}  & 4.5184e+0 (1.20e-1) $\approx$ & 2.9554e+0 (5.96e-1) $+$ & 4.7509e+0 (7.85e-3) $-$       & 4.6498e+0 (2.14e-2) $-$       & 2.5044e+0 (2.14e-1) $+$       & 4.5682e+0 (1.06e-1) $\approx$ \\
                                   & 10  & 5.7510e+0 (1.64e-1)      & 2.6812e+0 (3.90e-1) $+$       & 6.0207e+0 (7.32e-3) $-$       & 5.7885e+0 (9.12e-2) $\approx$ & 3.8026e+0 (7.27e-1) $+$ & \hl{6.0222e+0 (7.39e-3) $-$}  & 5.9551e+0 (2.29e-2) $-$       & 3.0058e+0 (2.59e-1) $+$       & 5.8556e+0 (1.26e-1) $-$       \\
                                   & 13  & 7.5198e+0 (1.59e-2)      & 2.3017e+0 (5.74e-1) $+$       & 7.5094e+0 (1.63e-2) $+$       & 7.2694e+0 (1.56e-1) $+$       & 3.1324e+0 (1.04e+0) $+$ & \hl{7.5435e+0 (1.28e-2) $-$}  & 7.3846e+0 (4.21e-2) $+$       & 3.2040e+0 (5.59e-1) $+$       & 7.3900e+0 (4.78e-2) $+$       \\
                                   & 15  & 9.2694e+0 (2.15e-2)      & 2.8204e+0 (8.21e-1) $+$       & 9.2620e+0 (2.24e-2) $\approx$ & 8.8887e+0 (2.38e-1) $+$       & 4.4246e+0 (1.11e+0) $+$ & \hl{9.3309e+0 (2.28e-2) $-$}  & 9.0945e+0 (4.73e-2) $+$       & 4.1989e+0 (5.92e-1) $+$       & 9.0861e+0 (5.69e-2) $+$       \\
                                   & 20  & 1.3691e+1 (8.08e-2)      & 4.1923e+0 (8.05e-1) $+$       & 1.3680e+1 (6.96e-2) $\approx$ & 1.3274e+1 (3.73e-1) $+$       & 4.8941e+0 (1.65e+0) $+$ & \hl{1.4097e+1 (5.86e-2) $-$}  & 1.3682e+1 (9.26e-2) $\approx$ & 5.6036e+0 (8.74e-1) $+$       & 1.3395e+1 (5.71e-2) $+$       \\
            \hline
            \multirow{6}{*}{MaF9}  & 5   & 4.2831e+0 (4.29e-1)      & 4.0467e+0 (1.68e-1) $+$       & \hl{4.8424e+0 (9.69e-2) $-$}  & 3.7268e+0 (5.37e-1) $+$       & 3.1527e+0 (8.71e-1) $+$ & 3.7156e+0 (6.58e-1) $+$       & 4.7642e+0 (6.70e-2) $-$       & 4.4039e+0 (8.71e-2) $\approx$ & 4.6544e+0 (1.89e-1) $-$       \\
                                   & 8   & 4.3374e+0 (9.52e-1)      & 3.4859e+0 (5.02e-1) $+$       & \hl{7.1994e+0 (2.12e-1) $-$}  & 3.1375e+0 (1.43e+0) $+$       & 3.6610e-1 (5.87e-1) $+$ & 6.3530e+0 (9.13e-1) $-$       & 6.5046e+0 (5.36e-1) $-$       & 2.9632e+0 (6.68e-1) $+$       & 4.8447e-1 (7.00e-1) $+$       \\
                                   & 10  & 7.7178e+0 (9.60e-1)      & 3.9554e+0 (7.52e-1) $+$       & \hl{9.6729e+0 (7.31e-2) $-$}  & 6.0615e+0 (1.07e+0) $+$       & 3.9807e-1 (8.14e-1) $+$ & 9.0052e+0 (9.59e-1) $-$       & 9.0899e+0 (9.63e-2) $-$       & 1.9736e+0 (5.51e-1) $+$       & 7.9332e-1 (4.87e-1) $+$       \\
                                   & 13  & 1.4875e+1 (6.59e-1)      & 3.9172e+0 (1.13e+0) $+$       & \hl{1.5372e+1 (5.09e-2) $-$}  & 9.7006e+0 (2.42e+0) $+$       & 5.5523e+0 (2.83e+0) $+$ & 1.1968e+1 (2.69e+0) $+$       & 1.4639e+1 (1.26e-1) $+$       & 6.0378e+0 (2.64e+0) $+$       & 1.0128e+1 (4.02e+0) $+$       \\
                                   & 15  & 1.8359e+1 (2.71e+0)      & 4.3222e+0 (1.27e+0) $+$       & \hl{1.9639e+1 (2.76e-2) $-$}  & 1.4563e+1 (1.22e+0) $+$       & 8.0749e+0 (4.34e+0) $+$ & 1.6596e+1 (2.96e+0) $+$       & 1.8730e+1 (1.17e-1) $-$       & 6.7414e+0 (4.03e+0) $+$       & 1.4313e+1 (5.41e+0) $+$       \\
                                   & 20  & 2.0191e+1 (8.93e+0)      & 4.3518e+0 (1.64e+0) $+$       & \hl{2.9161e+1 (5.41e+0) $-$}  & 8.8719e+0 (2.84e+0) $+$       & 8.3947e-1 (1.82e+0) $+$ & 2.7681e+1 (7.35e-1) $-$       & 2.6718e+1 (3.36e-1) $-$       & 9.5215e+0 (4.63e+0) $+$       & 2.1367e-1 (4.61e-1) $+$       \\
            \hline
            \multirow{6}{*}{MaF10} & 5   & 7.5854e+0 (2.96e-3)      & 7.5902e+0 (7.60e-4) $-$       & 7.5885e+0 (9.06e-4) $-$       & 7.5837e+0 (3.23e-3) $+$       & 6.3699e+0 (6.56e-1) $+$ & 7.5918e+0 (1.90e-4) $-$       & 7.5923e+0 (1.87e-4) $-$       & 7.5913e+0 (2.67e-4) $-$       & \hl{7.5924e+0 (2.64e-4) $-$}  \\
                                   & 8   & 2.5614e+1 (6.24e-3)      & 2.5608e+1 (4.86e-3) $+$       & 2.5615e+1 (6.37e-3) $\approx$ & 2.5564e+1 (4.18e-2) $+$       & 2.3220e+1 (1.56e+0) $+$ & \hl{2.5629e+1 (2.54e-5) $-$}  & 2.5629e+1 (5.41e-5) $-$       & 2.5629e+1 (3.34e-5) $-$       & 2.5629e+1 (2.40e-5) $-$       \\
                                   & 10  & 5.7633e+1 (1.23e-2)      & 5.7635e+1 (7.83e-3) $\approx$ & 5.7637e+1 (7.31e-3) $\approx$ & 5.7516e+1 (8.18e-2) $+$       & 5.3694e+1 (3.88e+0) $+$ & 5.7665e+1 (2.47e-5) $-$       & 5.7665e+1 (3.68e-5) $-$       & \hl{5.7665e+1 (1.81e-7) $-$}  & 5.7665e+1 (6.46e-6) $-$       \\
                                   & 13  & 1.9445e+2 (6.41e-2)      & 1.9452e+2 (4.32e-2) $-$       & 1.9449e+2 (4.02e-2) $-$       & 1.9386e+2 (2.68e-1) $+$       & 1.7931e+2 (1.19e+1) $+$ & \hl{1.9462e+2 (6.23e-14) $-$} & 1.9462e+2 (3.68e-4) $-$       & 1.9462e+2 (2.03e-4) $-$       & 1.9462e+2 (1.54e-4) $-$       \\
                                   & 15  & 4.3768e+2 (1.07e-1)      & 4.3769e+2 (1.03e-1) $\approx$ & 4.3770e+2 (6.96e-2) $\approx$ & 4.3653e+2 (5.06e-1) $+$       & 4.1688e+2 (2.78e+1) $+$ & \hl{4.3789e+2 (0.00e+0) $-$}  & 4.3789e+2 (2.81e-4) $-$       & 4.3789e+2 (3.41e-4) $-$       & 4.3789e+2 (4.25e-4) $-$       \\
                                   & 20  & 3.3243e+3 (4.74e-1)      & 3.3245e+3 (3.35e-1) $\approx$ & 3.3236e+3 (5.77e-1) $+$       & 3.3145e+3 (4.31e+0) $+$       & 3.2232e+3 (1.50e+2) $+$ & \hl{3.3253e+3 (0.00e+0) $-$}  & 3.3253e+3 (6.57e-4) $-$       & 3.3253e+3 (5.39e-5) $-$       & 3.3253e+3 (2.97e-4) $-$       \\
            \hline
            \multirow{6}{*}{MaF11} & 5   & 7.5405e+0 (9.24e-3)      & 7.5107e+0 (1.22e-2) $+$       & 7.5348e+0 (8.50e-3) $+$       & 7.4696e+0 (1.90e-2) $+$       & 7.3753e+0 (2.50e-2) $+$ & 7.5166e+0 (1.36e-2) $+$       & 7.5536e+0 (8.97e-3) $-$       & \hl{7.5661e+0 (1.29e-2) $-$}  & 7.5607e+0 (9.40e-3) $-$       \\
                                   & 8   & 2.5518e+1 (1.94e-2)      & 2.5200e+1 (9.32e-2) $+$       & 2.5478e+1 (2.35e-2) $+$       & 2.5291e+1 (8.96e-2) $+$       & 2.4584e+1 (1.46e-1) $+$ & 2.5419e+1 (3.68e-2) $+$       & 2.5466e+1 (5.14e-2) $+$       & \hl{2.5564e+1 (7.15e-2) $-$}  & 2.5494e+1 (4.04e-2) $+$       \\
                                   & 10  & 5.7442e+1 (4.80e-2)      & 5.6896e+1 (1.79e-1) $+$       & 5.7366e+1 (5.39e-2) $+$       & 5.7049e+1 (1.60e-1) $+$       & 5.4892e+1 (4.16e-1) $+$ & 5.7259e+1 (8.55e-2) $+$       & 5.7295e+1 (8.88e-2) $+$       & \hl{5.7585e+1 (4.12e-2) $-$}  & 5.7411e+1 (8.21e-2) $\approx$ \\
                                   & 13  & \hl{1.9416e+2 (1.06e-1)} & 1.9037e+2 (1.16e+0) $+$       & 1.9391e+2 (1.26e-1) $+$       & 1.9210e+2 (9.05e-1) $+$       & 1.8414e+2 (6.52e+0) $+$ & 1.9379e+2 (2.64e-1) $+$       & 1.9406e+2 (2.31e-1) $\approx$ & 1.9415e+2 (1.83e-1) $\approx$ & 1.9379e+2 (3.97e-1) $+$       \\
                                   & 15  & \hl{4.3693e+2 (2.09e-1)} & 4.2979e+2 (2.14e+0) $+$       & 4.3652e+2 (2.48e-1) $+$       & 4.3239e+2 (2.32e+0) $+$       & 4.1595e+2 (4.04e+0) $+$ & 4.3603e+2 (4.91e-1) $+$       & 4.3655e+2 (5.13e-1) $+$       & 4.3690e+2 (4.93e-1) $\approx$ & 4.3551e+2 (1.11e+0) $+$       \\
                                   & 20  & \hl{3.3210e+3 (1.35e+0)} & 3.2948e+3 (1.82e+1) $+$       & 3.3187e+3 (1.96e+0) $+$       & 3.2972e+3 (9.80e+0) $+$       & 3.1220e+3 (9.93e+1) $+$ & 3.3155e+3 (4.67e+0) $+$       & 3.3171e+3 (4.16e+0) $+$       & 3.3172e+3 (4.34e+0) $+$       & 3.3094e+3 (7.11e+0) $+$       \\
            \hline
            \multirow{6}{*}{MaF12} & 5   & \hl{7.1568e+0 (2.23e-2)} & 7.0827e+0 (4.44e-2) $+$       & 7.1420e+0 (2.40e-2) $+$       & 6.9357e+0 (3.97e-2) $+$       & 5.9794e+0 (2.70e-1) $+$ & 6.9764e+0 (9.17e-2) $+$       & 6.9639e+0 (6.54e-2) $+$       & 6.8121e+0 (3.35e-1) $+$       & 6.8130e+0 (1.92e-1) $+$       \\
                                   & 8   & \hl{2.4548e+1 (7.38e-2)} & 2.4178e+1 (2.31e-1) $+$       & 2.4443e+1 (1.28e-1) $+$       & 2.3761e+1 (2.05e-1) $+$       & 1.7587e+1 (1.19e+0) $+$ & 2.3911e+1 (8.91e-1) $+$       & 2.3527e+1 (4.11e-1) $+$       & 2.4365e+1 (8.05e-1) $+$       & 2.2333e+1 (5.70e-1) $+$       \\
                                   & 10  & \hl{5.5224e+1 (2.19e-1)} & 5.4321e+1 (1.37e+0) $+$       & 5.5045e+1 (2.51e-1) $+$       & 5.3938e+1 (7.79e-1) $+$       & 3.9537e+1 (3.39e+0) $+$ & 5.4461e+1 (1.14e+0) $+$       & 5.3493e+1 (6.59e-1) $+$       & 5.4339e+1 (2.66e+0) $\approx$ & 4.8343e+1 (1.31e+0) $+$       \\
                                   & 13  & \hl{1.8546e+2 (7.01e-1)} & 1.8065e+2 (5.81e+0) $+$       & 1.8432e+2 (4.25e+0) $\approx$ & 1.8323e+2 (4.33e+0) $+$       & 1.3294e+2 (1.19e+1) $+$ & 1.8248e+2 (7.45e+0) $\approx$ & 1.7576e+2 (5.86e+0) $+$       & 1.8119e+2 (5.03e+0) $+$       & 1.6361e+2 (5.58e+0) $+$       \\
                                   & 15  & \hl{4.1432e+2 (8.72e+0)} & 3.9103e+2 (1.75e+1) $+$       & 4.1403e+2 (1.01e+1) $\approx$ & 4.1274e+2 (1.50e+1) $\approx$ & 2.9906e+2 (3.07e+1) $+$ & 4.0797e+2 (2.13e+1) $\approx$ & 3.8730e+2 (1.53e+1) $+$       & 3.9608e+2 (1.78e+1) $+$       & 3.5285e+2 (1.08e+1) $+$       \\
                                   & 20  & 3.1300e+3 (1.03e+2)      & 2.9856e+3 (1.11e+2) $+$       & 3.1166e+3 (8.02e+1) $+$       & \hl{3.1447e+3 (1.81e+2) $-$}  & 2.1799e+3 (2.05e+2) $+$ & 3.1308e+3 (1.39e+2) $-$       & 2.9837e+3 (1.34e+2) $+$       & 3.0387e+3 (1.93e+2) $\approx$ & 2.6983e+3 (1.27e+2) $+$       \\
            \hline
            \multirow{6}{*}{MaF13} & 5   & \hl{4.5399e+0 (7.23e-2)} & 3.4830e+0 (2.58e-1) $+$       & 4.4765e+0 (7.89e-2) $+$       & 3.9941e+0 (1.77e-1) $+$       & 2.1651e+0 (8.63e-2) $+$ & 4.2007e+0 (1.30e-1) $+$       & 4.3949e+0 (5.65e-2) $+$       & 3.9249e+0 (6.98e-2) $+$       & 4.4634e+0 (7.72e-2) $+$       \\
                                   & 8   & \hl{1.0993e+1 (1.16e-1)} & 6.7615e+0 (1.52e+0) $+$       & 1.0862e+1 (1.77e-1) $+$       & 9.0115e+0 (1.09e+0) $+$       & 2.2771e+0 (9.22e-2) $+$ & 9.7621e+0 (5.49e-1) $+$       & 1.0564e+1 (1.53e-1) $+$       & 8.9446e+0 (3.57e-1) $+$       & 1.0697e+1 (1.78e-1) $+$       \\
                                   & 10  & \hl{2.1161e+1 (1.62e-1)} & 1.4708e+1 (1.14e+0) $+$       & 2.0906e+1 (2.13e-1) $+$       & 1.7619e+1 (1.35e+0) $+$       & 2.5531e+0 (2.03e-1) $+$ & 1.8889e+1 (1.18e+0) $+$       & 2.0300e+1 (2.99e-1) $+$       & 1.6677e+1 (1.10e+0) $+$       & 2.0445e+1 (3.87e-1) $+$       \\
                                   & 13  & \hl{5.7863e+1 (2.70e-1)} & 3.8060e+1 (9.47e+0) $+$       & 5.7587e+1 (4.52e-1) $+$       & 3.8544e+1 (1.31e+1) $+$       & 2.8787e+0 (6.32e-2) $+$ & 4.8295e+1 (3.97e+0) $+$       & 5.5718e+1 (1.10e+0) $+$       & 4.1801e+1 (5.67e+0) $+$       & 5.5262e+1 (1.85e+0) $+$       \\
                                   & 15  & \hl{1.1735e+2 (4.38e-1)} & 8.2965e+1 (1.05e+1) $+$       & 1.1669e+2 (6.61e-1) $+$       & 8.9865e+1 (2.28e+1) $+$       & 3.3398e+0 (4.18e-1) $+$ & 1.0041e+2 (7.46e+0) $+$       & 1.1296e+2 (2.55e+0) $+$       & 8.2088e+1 (1.83e+1) $+$       & 1.1187e+2 (3.36e+0) $+$       \\
                                   & 20  & \hl{6.8891e+2 (4.31e+0)} & 4.9688e+2 (1.18e+2) $+$       & 6.8830e+2 (3.91e+0) $\approx$ & 4.4887e+2 (1.64e+2) $+$       & 4.5689e+0 (8.75e-2) $+$ & 5.8731e+2 (5.92e+1) $+$       & 6.6819e+2 (1.58e+1) $+$       & 3.6029e+2 (1.37e+2) $+$       & 6.5253e+2 (3.08e+1) $+$       \\
            \bottomrule
        \end{tabular}
    }
    \\
	\vspace{1mm}
	\footnotesize \raggedright \hspace{4mm}Each column represents each algorithm, and each row represents each test instance.
	
	\hspace{4mm}Each cell shows the median and standard deviation of the evaluations of HV and IGD${}^+$ in 31 runs.
	
	\hspace{4mm}The algorithm with the best median value for each row is highlighted.
	
	\hspace{4mm}The result of comparison with RVEA-CA at a 0.05 level by Wilcoxon's rank-sum test is shown on the right end of each cell in each row except for RVEA-CA.
	
	\hspace{4mm}``$+$'', ``$\approx$'', and ``$-$'' mean that RVEA-CA performs significantly better than, non-significantly better/worse than, and significantly worse than each algorithm, respectively.
    \label{tbl:HV2}
\end{sidewaystable*}

\begin{sidewaystable*}[htbp]
    \centering
    \caption{Median value and standard deviation of 31 runs on IGD${}^+$ (I)}
    \scalebox{0.66}{
        \begin{tabular}{ccccccccccc}
            \toprule
            Problem               & $M$ & RVEA-CA                  & RVEA                          & RVEA-iGNG                          & DEA-GNG                       & A-NSGA-III                    & VaEA                          & AR-MOEA                       & MOEA/D-AWA                    & AdaW                               \\
            \midrule
            \multirow{6}{*}{MaF1} & 5   & 7.6217e-2 (7.55e-4)      & 1.8898e-1 (1.02e-2) $+$       & 7.5822e-2 (5.41e-4) $\approx$      & 7.3609e-2 (2.03e-3) $-$       & 1.3354e-1 (1.61e-2) $+$       & 8.1887e-2 (1.05e-3) $+$       & 8.4430e-2 (1.34e-3) $+$       & 1.2052e-1 (1.45e-3) $+$       & \hl{7.2048e-2 (3.07e-4) $-$}       \\
                                  & 8   & 1.4685e-1 (1.33e-3)      & 3.5573e-1 (2.61e-2) $+$       & 1.4861e-1 (2.08e-3) $+$            & 1.4683e-1 (4.50e-3) $\approx$ & 2.1815e-1 (9.85e-3) $+$       & 1.4182e-1 (9.24e-4) $-$       & 1.4468e-1 (1.21e-3) $-$       & 3.0247e-1 (2.40e-2) $+$       & \hl{1.3599e-1 (5.37e-4) $-$}       \\
                                  & 10  & 1.7458e-1 (2.58e-3)      & 3.7536e-1 (2.22e-2) $+$       & 1.9632e-1 (3.36e-2) $+$            & 1.6571e-1 (3.85e-3) $-$       & 2.2709e-1 (7.35e-3) $+$       & 1.5976e-1 (1.12e-3) $-$       & 1.6302e-1 (2.62e-3) $-$       & 3.1285e-1 (2.88e-2) $+$       & \hl{1.5908e-1 (9.94e-4) $-$}       \\
                                  & 13  & 2.4428e-1 (1.00e-2)      & 4.3387e-1 (1.50e-2) $+$       & 3.0533e-1 (3.15e-2) $+$            & 2.3066e-1 (1.02e-2) $-$       & 2.9721e-1 (8.28e-3) $+$       & 2.0893e-1 (1.55e-3) $-$       & 2.1443e-1 (3.10e-3) $-$       & 3.5036e-1 (4.47e-2) $+$       & \hl{2.0831e-1 (1.50e-3) $-$}       \\
                                  & 15  & 2.3518e-1 (5.77e-3)      & 4.0704e-1 (1.38e-2) $+$       & 3.0921e-1 (2.25e-2) $+$            & 2.1668e-1 (8.39e-3) $-$       & 2.7516e-1 (1.05e-2) $+$       & 2.0083e-1 (1.45e-3) $-$       & 2.1020e-1 (2.61e-3) $-$       & 3.1075e-1 (3.77e-2) $+$       & \hl{1.9685e-1 (1.54e-3) $-$}       \\
                                  & 20  & 3.5706e-1 (2.22e-2)      & 5.0681e-1 (7.47e-3) $+$       & 4.0675e-1 (2.71e-2) $+$            & 3.3911e-1 (2.68e-2) $-$       & 4.1029e-1 (1.35e-2) $+$       & 2.9419e-1 (8.10e-4) $-$       & 2.8685e-1 (5.21e-3) $-$       & 4.4996e-1 (2.95e-2) $+$       & \hl{2.8129e-1 (1.97e-3) $-$}       \\
            \hline
            \multirow{6}{*}{MaF2} & 5   & 8.6830e-2 (1.83e-3)      & 1.2276e-1 (1.92e-3) $+$       & 8.7277e-2 (1.80e-3) $\approx$      & \hl{8.2639e-2 (2.97e-3) $-$}  & 1.9003e-1 (3.58e-2) $+$       & 1.0260e-1 (2.43e-3) $+$       & 1.0898e-1 (2.07e-3) $+$       & 1.4019e-1 (4.11e-3) $+$       & 9.2800e-2 (1.92e-3) $+$            \\
                                  & 8   & 1.4545e-1 (5.25e-3)      & 3.0101e-1 (8.47e-2) $+$       & \hl{1.4415e-1 (3.00e-3) $\approx$} & 1.9797e-1 (2.07e-2) $+$       & 3.3456e-1 (9.78e-2) $+$       & 1.7870e-1 (2.20e-3) $+$       & 1.8530e-1 (6.59e-3) $+$       & 2.1011e-1 (1.09e-2) $+$       & 1.5455e-1 (4.44e-3) $+$            \\
                                  & 10  & 1.7680e-1 (4.43e-3)      & 3.1161e-1 (1.03e-1) $+$       & \hl{1.7599e-1 (3.39e-3) $\approx$} & 2.3633e-1 (1.75e-2) $+$       & 3.7788e-1 (8.91e-2) $+$       & 2.1402e-1 (3.40e-3) $+$       & 2.0720e-1 (5.41e-3) $+$       & 2.2999e-1 (1.13e-2) $+$       & 1.9112e-1 (5.31e-3) $+$            \\
                                  & 13  & 2.3148e-1 (8.33e-3)      & 6.5350e-1 (1.88e-1) $+$       & \hl{2.2454e-1 (5.58e-3) $-$}       & 2.7862e-1 (1.91e-2) $+$       & 5.9223e-1 (2.05e-1) $+$       & 2.5056e-1 (3.15e-3) $+$       & 2.6739e-1 (1.13e-2) $+$       & 3.0746e-1 (2.48e-2) $+$       & 2.3604e-1 (5.96e-3) $+$            \\
                                  & 15  & \hl{2.2946e-1 (8.17e-3)} & 6.4419e-1 (2.02e-1) $+$       & 2.3154e-1 (7.42e-3) $\approx$      & 2.6708e-1 (1.40e-2) $+$       & 4.6951e-1 (9.70e-2) $+$       & 2.6049e-1 (2.83e-3) $+$       & 2.7468e-1 (7.80e-3) $+$       & 2.9524e-1 (1.81e-2) $+$       & 2.4343e-1 (4.81e-3) $+$            \\
                                  & 20  & \hl{2.4695e-1 (1.01e-2)} & 7.6663e-1 (2.15e-1) $+$       & 2.5777e-1 (8.63e-3) $+$            & 2.8732e-1 (3.02e-2) $+$       & 4.8900e-1 (5.97e-2) $+$       & 2.8782e-1 (3.10e-3) $+$       & 2.5372e-1 (1.49e-2) $+$       & 2.7741e-1 (1.96e-2) $+$       & 2.7615e-1 (5.41e-3) $+$            \\
            \hline
            \multirow{6}{*}{MaF3} & 5   & 1.2051e+1 (2.37e+1)      & 4.1232e+1 (3.13e+1) $+$       & 3.1790e-1 (8.39e-1) $-$            & 8.4175e+0 (1.32e+1) $\approx$ & 3.3692e+0 (4.28e+0) $\approx$ & 7.8167e+0 (2.54e+1) $\approx$ & 7.3281e-1 (1.85e+0) $-$       & \hl{6.8448e-2 (3.12e-2) $-$}  & 1.0315e-1 (1.85e-1) $-$            \\
                                  & 8   & 7.3363e+0 (1.18e+1)      & 2.9664e+0 (6.02e+0) $\approx$ & 1.6175e+0 (5.23e+0) $-$            & 7.0789e+1 (1.99e+2) $+$       & 4.2494e+3 (4.30e+3) $+$       & 7.6594e+3 (1.89e+4) $+$       & 1.5288e-1 (7.28e-1) $-$       & \hl{2.1892e-2 (1.45e-3) $-$}  & 6.9603e-2 (2.26e-2) $-$            \\
                                  & 10  & 6.9026e+0 (2.35e+1)      & 2.3416e+0 (7.55e+0) $\approx$ & 2.7619e+0 (6.23e+0) $\approx$      & 3.2837e+1 (5.78e+1) $+$       & 4.3213e+3 (2.72e+3) $+$       & 5.4003e+4 (1.03e+5) $+$       & 1.7921e-1 (8.76e-1) $-$       & \hl{2.4873e-2 (8.65e-3) $-$}  & 7.2959e-2 (3.94e-2) $-$            \\
                                  & 13  & 1.6861e-2 (2.45e-3)      & \hl{1.3054e-2 (1.76e-3) $-$}  & 2.5454e-2 (6.14e-2) $+$            & 1.4123e+1 (2.94e+1) $+$       & 2.5238e+3 (2.45e+3) $+$       & 3.6010e+3 (6.19e+3) $+$       & 9.8604e-1 (3.04e+0) $\approx$ & 1.8824e-2 (6.24e-3) $\approx$ & 5.3460e-2 (2.25e-2) $+$            \\
                                  & 15  & 2.0420e-2 (4.70e-3)      & 2.0403e-2 (4.68e-2) $-$       & \hl{1.6030e-2 (1.37e-2) $-$}       & 1.9496e+0 (5.34e+0) $+$       & 3.2443e+3 (2.75e+3) $+$       & 4.8359e+3 (9.28e+3) $+$       & 4.1218e+0 (7.52e+0) $\approx$ & 1.6282e-2 (4.11e-3) $-$       & 6.6580e-2 (4.85e-2) $+$            \\
                                  & 20  & 1.4520e-2 (3.88e-3)      & \hl{6.2341e-3 (1.08e-3) $-$}  & 1.1433e-2 (4.23e-3) $-$            & 8.9719e-1 (3.79e+0) $+$       & 3.1878e+7 (8.10e+7) $+$       & 2.0540e+3 (4.57e+3) $+$       & 1.2590e+0 (3.63e+0) $\approx$ & 2.2748e-2 (6.25e-3) $+$       & 6.1094e-2 (3.71e-2) $+$            \\
            \hline
            \multirow{6}{*}{MaF4} & 5   & 4.6272e-1 (7.44e-1)      & 6.0360e-1 (8.75e-1) $+$       & 1.7560e-1 (2.22e-1) $-$            & 1.5673e-1 (1.04e-1) $-$       & 1.1106e+0 (1.61e+0) $+$       & 1.7985e-1 (6.81e-2) $\approx$ & 1.9437e-1 (2.14e-1) $-$       & 3.1295e-1 (1.21e-2) $-$       & \hl{1.3692e-1 (1.96e-2) $-$}       \\
                                  & 8   & 2.2399e-1 (3.46e-2)      & 7.4760e-1 (3.77e-1) $+$       & 2.4788e-1 (6.15e-2) $+$            & 2.2817e-1 (2.41e-2) $\approx$ & 1.1265e+0 (1.88e+0) $+$       & 2.4191e-1 (2.49e-2) $+$       & 4.3866e-1 (2.97e-2) $+$       & 5.0989e-1 (7.77e-2) $+$       & \hl{2.1015e-1 (1.93e-2) $\approx$} \\
                                  & 10  & \hl{2.6733e-1 (2.66e-2)} & 7.3515e-1 (4.02e-2) $+$       & 3.2004e-1 (6.24e-2) $+$            & 2.6882e-1 (2.66e-2) $\approx$ & 8.5065e-1 (1.80e+0) $+$       & 2.7577e-1 (1.62e-2) $+$       & 5.8027e-1 (2.89e-2) $+$       & 5.4146e-1 (8.12e-2) $+$       & 2.7178e-1 (2.46e-2) $\approx$      \\
                                  & 13  & \hl{3.1884e-1 (1.28e-2)} & 8.2280e-1 (5.11e-2) $+$       & 3.5454e-1 (2.05e-2) $+$            & 3.6757e-1 (2.01e-2) $+$       & 5.5084e-1 (5.76e-1) $+$       & 3.1929e-1 (9.53e-3) $\approx$ & 7.0941e-1 (1.43e-2) $+$       & 5.5182e-1 (9.89e-2) $+$       & 3.2488e-1 (1.87e-2) $\approx$      \\
                                  & 15  & 3.8754e-1 (6.23e-2)      & 8.3809e-1 (4.66e-2) $+$       & 4.1991e-1 (4.58e-2) $+$            & 3.7813e-1 (2.27e-2) $\approx$ & 4.6368e-1 (5.08e-2) $+$       & \hl{3.3119e-1 (2.23e-2) $-$}  & 7.4102e-1 (1.97e-2) $+$       & 5.1705e-1 (7.50e-2) $+$       & 3.4394e-1 (2.12e-2) $-$            \\
                                  & 20  & 5.0095e-1 (6.77e-2)      & 8.7951e-1 (2.59e-2) $+$       & 5.6205e-1 (4.72e-2) $+$            & 4.6558e-1 (1.89e-2) $\approx$ & 1.1772e+0 (2.78e+0) $+$       & \hl{3.7022e-1 (1.42e-2) $-$}  & 8.1259e-1 (1.15e-2) $+$       & 7.2915e-1 (7.51e-2) $+$       & 3.9110e-1 (1.10e-2) $-$            \\
            \hline
            \multirow{6}{*}{MaF5} & 5   & 7.3935e-2 (1.49e-2)      & 7.0979e-2 (3.85e-2) $-$       & 7.8469e-2 (1.99e-2) $+$            & 8.4367e-2 (1.85e-3) $+$       & 1.1199e-1 (2.12e-2) $+$       & 8.4446e-2 (2.43e-3) $+$       & \hl{6.4172e-2 (2.94e-4) $-$}  & 1.5353e-1 (9.51e-2) $+$       & 7.9929e-2 (2.43e-2) $\approx$      \\
                                  & 8   & 1.4170e-1 (2.93e-3)      & 1.4906e-1 (1.62e-2) $+$       & 1.5249e-1 (1.64e-2) $+$            & 1.6213e-1 (4.64e-3) $+$       & 3.2759e-1 (5.16e-2) $+$       & 1.7465e-1 (5.89e-3) $+$       & \hl{1.2502e-1 (9.22e-4) $-$}  & 2.2261e-1 (4.61e-2) $+$       & 1.5316e-1 (6.36e-3) $+$            \\
                                  & 10  & 1.8471e-1 (3.04e-3)      & 1.9120e-1 (6.00e-3) $+$       & 1.8947e-1 (6.74e-3) $+$            & 1.9436e-1 (3.85e-3) $+$       & 4.9500e-1 (5.29e-2) $+$       & 2.3159e-1 (8.55e-3) $+$       & \hl{1.8167e-1 (7.01e-4) $-$}  & 2.6128e-1 (3.88e-2) $+$       & 1.8855e-1 (5.78e-3) $+$            \\
                                  & 13  & 2.3350e-1 (4.56e-3)      & 2.7676e-1 (3.21e-2) $+$       & 2.6409e-1 (1.38e-2) $+$            & 2.3366e-1 (6.35e-3) $\approx$ & 6.7618e-1 (5.41e-2) $+$       & 2.4392e-1 (5.20e-3) $+$       & \hl{2.2669e-1 (1.90e-3) $-$}  & 3.3707e-1 (5.22e-2) $+$       & 2.4663e-1 (7.06e-3) $+$            \\
                                  & 15  & 2.5654e-1 (5.45e-3)      & 2.7337e-1 (8.64e-3) $+$       & 2.8636e-1 (8.43e-3) $+$            & \hl{2.3509e-1 (3.78e-3) $-$}  & 6.5567e-1 (7.34e-2) $+$       & 2.8307e-1 (1.41e-2) $+$       & 2.4485e-1 (1.98e-3) $-$       & 3.4471e-1 (4.37e-2) $+$       & 2.5509e-1 (8.06e-3) $\approx$      \\
                                  & 20  & 3.2201e-1 (7.61e-3)      & 3.1175e-1 (1.48e-2) $-$       & 3.4319e-1 (1.23e-2) $+$            & \hl{2.4544e-1 (4.08e-3) $-$}  & 7.8558e-1 (8.17e-2) $+$       & 3.6005e-1 (2.01e-2) $+$       & 2.4639e-1 (4.83e-3) $-$       & 3.7200e-1 (3.11e-2) $+$       & 2.9629e-1 (9.35e-3) $-$            \\
            \hline
            \multirow{6}{*}{MaF6} & 5   & 1.2324e-3 (2.45e-5)      & 4.2037e-2 (6.81e-3) $+$       & \hl{1.2237e-3 (2.27e-5) $\approx$} & 1.2393e-3 (4.48e-5) $\approx$ & 8.8481e-3 (7.17e-3) $+$       & 1.6571e-3 (5.60e-5) $+$       & 1.2670e-3 (2.95e-5) $+$       & 2.1116e-2 (3.18e-3) $+$       & 1.2947e-3 (3.10e-5) $+$            \\
                                  & 8   & 3.2663e-1 (6.45e-1)      & 1.2361e-1 (5.46e-2) $-$       & \hl{1.1309e-3 (1.37e-5) $-$}       & 7.6602e-1 (1.70e+0) $+$       & 6.8563e-1 (1.23e+0) $+$       & 3.7618e-1 (7.51e-1) $+$       & 1.2223e-3 (2.39e-5) $-$       & 2.1709e-2 (1.48e-2) $-$       & 1.3015e-3 (1.85e-5) $-$            \\
                                  & 10  & 1.9328e+0 (1.50e+0)      & 7.0109e-2 (1.93e-2) $-$       & \hl{1.0306e-3 (8.93e-6) $-$}       & 2.3527e+0 (1.68e+0) $\approx$ & 1.9190e+0 (9.08e-1) $\approx$ & 1.4280e+0 (6.22e-1) $\approx$ & 1.2793e-1 (2.00e-1) $-$       & 2.1524e-2 (1.39e-2) $-$       & 1.2272e-3 (1.56e-5) $-$            \\
                                  & 13  & 1.3662e+0 (9.43e-1)      & 2.4439e-1 (1.33e-1) $-$       & 5.8426e-2 (3.16e-1) $-$            & 3.6590e+0 (8.55e+0) $\approx$ & 7.1941e-1 (2.79e-1) $-$       & 4.7235e-1 (2.80e-1) $-$       & 6.6914e-2 (5.62e-2) $-$       & 1.9871e-2 (1.62e-2) $-$       & \hl{2.0095e-3 (2.50e-5) $-$}       \\
                                  & 15  & 2.0996e+0 (9.81e-1)      & 1.8690e-1 (1.20e-1) $-$       & 1.5213e-1 (5.60e-1) $-$            & 1.0473e+1 (2.03e+1) $\approx$ & 7.5673e-1 (2.79e-1) $-$       & 6.2769e-1 (2.36e-1) $-$       & 9.3251e-2 (7.22e-2) $-$       & 1.1487e-2 (6.35e-3) $-$       & \hl{1.6042e-3 (1.46e-5) $-$}       \\
                                  & 20  & 9.8771e+0 (1.99e+1)      & 9.7302e-2 (9.34e-7) $-$       & 1.3314e+1 (2.10e+1) $\approx$      & 2.1054e+1 (2.17e+1) $\approx$ & 1.0519e+1 (8.90e+0) $+$       & 3.8876e-1 (8.64e-2) $-$       & 7.2228e-2 (3.44e-2) $-$       & 1.8299e-2 (1.47e-2) $-$       & \hl{1.8387e-3 (2.00e-5) $-$}       \\
            \hline
            \multirow{6}{*}{MaF7} & 5   & 2.2223e-1 (2.23e-2)      & 1.3555e-1 (4.02e-3) $-$       & 7.9482e-2 (3.82e-3) $-$            & \hl{7.9393e-2 (7.66e-3) $-$}  & 1.6701e-1 (2.47e-2) $-$       & 1.1712e-1 (3.88e-3) $-$       & 9.8469e-2 (2.30e-3) $-$       & 1.4897e-1 (2.17e-2) $-$       & 9.3448e-2 (6.21e-3) $-$            \\
                                  & 8   & \hl{1.4712e-1 (6.88e-3)} & 2.5186e-1 (1.53e-2) $+$       & 1.5080e-1 (4.67e-3) $+$            & 1.7904e-1 (6.21e-3) $+$       & 3.4145e-1 (1.97e-2) $+$       & 2.1076e-1 (6.04e-3) $+$       & 2.1401e-1 (1.01e-2) $+$       & 2.1207e-1 (5.01e-3) $+$       & 2.0433e-1 (9.53e-3) $+$            \\
                                  & 10  & \hl{1.7149e-1 (9.43e-3)} & 3.2105e-1 (2.34e-2) $+$       & 1.8218e-1 (7.20e-3) $+$            & 2.2787e-1 (7.19e-3) $+$       & 4.2053e-1 (4.73e-2) $+$       & 2.3327e-1 (7.19e-3) $+$       & 3.0009e-1 (1.59e-2) $+$       & 2.1682e-1 (1.04e-2) $+$       & 2.2921e-1 (1.83e-2) $+$            \\
                                  & 13  & 2.8944e-1 (2.19e-2)      & 4.0985e-1 (1.00e-1) $+$       & 3.0995e-1 (2.55e-2) $+$            & 3.1692e-1 (1.63e-2) $+$       & 1.5656e+0 (2.36e-1) $+$       & \hl{2.5573e-1 (6.17e-3) $-$}  & 4.1389e-1 (4.86e-2) $+$       & 5.8171e-1 (1.73e-1) $+$       & 3.2602e-1 (5.14e-2) $+$            \\
                                  & 15  & 3.5581e-1 (2.96e-2)      & 5.9991e-1 (7.36e-2) $+$       & 4.3592e-1 (2.85e-2) $+$            & 4.1792e-1 (2.08e-2) $+$       & 2.1816e+0 (2.35e-1) $+$       & \hl{2.8533e-1 (7.91e-3) $-$}  & 5.6370e-1 (1.22e-1) $+$       & 8.3643e-1 (2.79e-1) $+$       & 4.3309e-1 (7.13e-2) $+$            \\
                                  & 20  & 4.4949e-1 (5.74e-2)      & 4.5849e-1 (7.92e-3) $+$       & 6.0642e-1 (8.78e-2) $+$            & 4.4213e-1 (1.42e-2) $\approx$ & 2.8200e+0 (3.26e-1) $+$       & \hl{3.4530e-1 (8.97e-3) $-$}  & 6.6738e-1 (1.87e-1) $+$       & 4.5010e-1 (1.84e-2) $+$       & 1.8332e+0 (3.29e-1) $+$            \\
            \bottomrule
        \end{tabular}
    }
    \\
	\vspace{1mm}
	\footnotesize \raggedright \hspace{4mm}Each column represents each algorithm, and each row represents each test instance.
	
	\hspace{4mm}Each cell shows the median and standard deviation of the evaluations of HV and IGD${}^+$ in 31 runs.
	
	\hspace{4mm}The algorithm with the best median value for each row is highlighted.
	
	\hspace{4mm}The result of comparison with RVEA-CA at a 0.05 level by Wilcoxon's rank-sum test is shown on the right end of each cell in each row except for RVEA-CA.
	
	\hspace{4mm}``$+$'', ``$\approx$'', and ``$-$'' mean that RVEA-CA performs significantly better than, non-significantly better/worse than, and significantly worse than each algorithm, respectively.
    \label{tbl:IGDP1}
\end{sidewaystable*}

\begin{sidewaystable*}[htbp]
    \centering
    \caption{Median value and standard deviation of 31 runs on IGD${}^+$ (II)}
    \scalebox{0.66}{
        \begin{tabular}{ccccccccccc}
            \toprule
            Problem                & $M$ & RVEA-CA                  & RVEA                          & RVEA-iGNG                          & DEA-GNG                       & A-NSGA-III              & VaEA                          & AR-MOEA                            & MOEA/D-AWA                    & AdaW                               \\
            \midrule
            \multirow{6}{*}{MaF8}  & 5   & 6.6987e-2 (1.54e-2)      & 1.1200e-1 (9.68e-3) $+$       & \hl{2.4911e-2 (1.44e-3) $-$}       & 3.6744e-2 (7.42e-3) $-$       & 1.6648e-1 (6.62e-2) $+$ & 2.9020e-2 (5.97e-3) $-$       & 2.9891e-2 (1.49e-3) $-$            & 8.1661e-2 (9.98e-3) $+$       & 6.3750e-2 (3.01e-2) $\approx$      \\
                                   & 8   & 4.4458e-2 (9.17e-3)      & 2.4116e-1 (4.53e-2) $+$       & \hl{2.8917e-2 (2.62e-4) $-$}       & 4.2212e-2 (7.50e-3) $\approx$ & 1.9042e-1 (7.71e-2) $+$ & 3.2531e-2 (8.09e-4) $-$       & 3.6940e-2 (1.28e-3) $-$            & 2.2435e-1 (2.57e-2) $+$       & 4.2741e-2 (8.03e-3) $\approx$      \\
                                   & 10  & 4.5242e-2 (9.64e-3)      & 2.6954e-1 (4.40e-2) $+$       & \hl{3.0456e-2 (2.05e-4) $-$}       & 4.0686e-2 (4.72e-3) $\approx$ & 1.8350e-1 (6.86e-2) $+$ & 3.4326e-2 (5.75e-4) $-$       & 3.7501e-2 (1.23e-3) $-$            & 2.5262e-1 (2.93e-2) $+$       & 3.8545e-2 (7.65e-3) $-$            \\
                                   & 13  & \hl{4.4133e-2 (4.69e-4)} & 3.9441e-1 (7.55e-2) $+$       & 4.4360e-2 (4.08e-4) $\approx$      & 5.3290e-2 (5.28e-3) $+$       & 3.5573e-1 (1.16e-1) $+$ & 5.1481e-2 (5.78e-4) $+$       & 5.6327e-2 (2.28e-3) $+$            & 3.2707e-1 (5.33e-2) $+$       & 4.8483e-2 (1.58e-3) $+$            \\
                                   & 15  & \hl{4.1065e-2 (3.40e-4)} & 3.9598e-1 (8.73e-2) $+$       & 4.1110e-2 (2.53e-4) $\approx$      & 5.0780e-2 (5.67e-3) $+$       & 3.0170e-1 (9.35e-2) $+$ & 4.8450e-2 (9.37e-4) $+$       & 5.2951e-2 (2.04e-3) $+$            & 3.0612e-1 (5.38e-2) $+$       & 4.4980e-2 (1.27e-3) $+$            \\
                                   & 20  & 4.8918e-2 (4.73e-4)      & 3.9675e-1 (5.82e-2) $+$       & \hl{4.8810e-2 (3.81e-4) $\approx$} & 5.8254e-2 (5.24e-3) $+$       & 3.9918e-1 (1.11e-1) $+$ & 5.5281e-2 (1.20e-3) $+$       & 6.0555e-2 (1.59e-3) $+$            & 3.4839e-1 (6.01e-2) $+$       & 5.2434e-2 (5.02e-4) $+$            \\
            \hline
            \multirow{6}{*}{MaF9}  & 5   & 7.9712e-2 (4.33e-2)      & 1.1840e-1 (1.55e-2) $+$       & \hl{3.2148e-2 (1.22e-2) $-$}       & 1.7207e-1 (8.85e-2) $+$       & 2.2739e-1 (1.49e-1) $+$ & 1.7498e-1 (9.94e-2) $+$       & 3.6526e-2 (5.94e-3) $-$            & 9.7078e-2 (8.72e-3) $+$       & 4.9891e-2 (2.19e-2) $-$            \\
                                   & 8   & 2.0521e-1 (6.92e-2)      & 2.5695e-1 (4.41e-2) $+$       & \hl{4.4558e-2 (1.36e-2) $-$}       & 4.4512e-1 (4.72e-1) $+$       & 1.3801e+1 (3.05e+1) $+$ & 9.1523e-2 (6.60e-2) $-$       & 7.6379e-2 (3.46e-2) $-$            & 3.1877e-1 (6.65e-2) $+$       & 1.4464e+0 (6.89e-1) $+$            \\
                                   & 10  & 1.1210e-1 (3.90e-2)      & 3.0899e-1 (7.90e-2) $+$       & \hl{4.5112e-2 (2.13e-3) $-$}       & 1.9242e-1 (5.41e-2) $+$       & 7.0847e+0 (6.45e+0) $+$ & 7.3216e-2 (4.40e-2) $-$       & 6.0330e-2 (2.41e-3) $-$            & 5.3530e-1 (2.73e-1) $+$       & 7.6726e+0 (1.61e+1) $+$            \\
                                   & 13  & 5.6409e-2 (8.64e-3)      & 4.6007e-1 (1.22e-1) $+$       & \hl{5.2859e-2 (1.55e-3) $\approx$} & 1.8934e-1 (1.10e-1) $+$       & 9.2587e-1 (1.54e+0) $+$ & 1.4556e-1 (9.26e-2) $+$       & 5.7965e-2 (1.66e-3) $+$            & 8.2943e-1 (1.44e+0) $+$       & 4.6538e-1 (1.08e+0) $+$            \\
                                   & 15  & 6.6919e-2 (6.14e-2)      & 4.8651e-1 (1.08e-1) $+$       & \hl{4.8008e-2 (6.58e-4) $\approx$} & 1.2108e-1 (2.49e-2) $+$       & 1.4397e+0 (2.45e+0) $+$ & 1.0833e-1 (7.34e-2) $+$       & 5.4025e-2 (1.17e-3) $-$            & 1.5832e+0 (2.43e+0) $+$       & 5.2479e-1 (1.48e+0) $+$            \\
                                   & 20  & 2.0907e-1 (2.16e-1)      & 6.3931e-1 (1.18e-1) $+$       & 3.1484e-1 (1.42e+0) $+$            & 4.2672e-1 (1.33e-1) $+$       & 1.0404e+1 (7.30e+0) $+$ & \hl{7.6726e-2 (7.30e-3) $-$}  & 8.0609e-2 (2.79e-3) $-$            & 1.5537e+0 (2.86e+0) $+$       & 1.0887e+1 (7.98e+0) $+$            \\
            \hline
            \multirow{6}{*}{MaF10} & 5   & \hl{2.7156e-2 (4.14e-4)} & 3.0826e-2 (3.80e-4) $+$       & 2.7727e-2 (4.69e-4) $+$            & 2.8731e-2 (8.61e-4) $+$       & 5.5191e-1 (6.70e-2) $+$ & 3.0059e-2 (5.36e-4) $+$       & 3.2998e-2 (6.27e-4) $+$            & 5.4737e-2 (1.01e-3) $+$       & 4.5546e-2 (5.96e-3) $+$            \\
                                   & 8   & \hl{2.9814e-2 (1.06e-3)} & 3.0305e-2 (6.98e-4) $+$       & 3.0036e-2 (1.27e-3) $\approx$      & 3.1968e-2 (1.44e-3) $+$       & 5.1724e-1 (4.70e-2) $+$ & 3.0547e-2 (5.72e-4) $+$       & 3.0853e-2 (8.55e-4) $+$            & 3.9072e-2 (1.21e-3) $+$       & 7.8914e-2 (6.84e-3) $+$            \\
                                   & 10  & 3.2931e-2 (2.21e-3)      & 3.2886e-2 (1.53e-3) $\approx$ & 3.1972e-2 (1.61e-3) $-$            & 3.0536e-2 (2.20e-3) $-$       & 5.0638e-1 (5.80e-2) $+$ & \hl{2.6860e-2 (6.13e-4) $-$}  & 3.2989e-2 (8.37e-4) $\approx$      & 3.9895e-2 (2.73e-3) $+$       & 1.1678e-1 (4.87e-3) $+$            \\
                                   & 13  & 4.0521e-2 (3.11e-3)      & 2.6207e-2 (1.37e-3) $-$       & 3.8478e-2 (3.53e-3) $-$            & 3.4725e-2 (2.97e-3) $-$       & 4.4821e-1 (6.72e-2) $+$ & \hl{2.5206e-2 (8.77e-4) $-$}  & 2.6285e-2 (8.20e-4) $-$            & 3.1824e-2 (1.90e-3) $-$       & 1.7066e-1 (1.04e-2) $+$            \\
                                   & 15  & 4.0496e-2 (4.03e-3)      & 2.4085e-2 (1.69e-3) $-$       & 3.9535e-2 (3.66e-3) $\approx$      & 3.2080e-2 (3.11e-3) $-$       & 4.2554e-1 (7.58e-2) $+$ & \hl{1.8182e-2 (5.82e-4) $-$}  & 2.4187e-2 (6.43e-4) $-$            & 2.8307e-2 (1.93e-3) $-$       & 1.7392e-1 (1.24e-2) $+$            \\
                                   & 20  & 3.0192e-2 (1.77e-3)      & 3.1354e-2 (1.92e-3) $+$       & 2.9282e-2 (1.86e-3) $-$            & 3.0573e-2 (2.05e-3) $\approx$ & 3.5397e-1 (6.28e-2) $+$ & \hl{1.8272e-2 (2.87e-4) $-$}  & 2.8997e-2 (1.19e-3) $-$            & 5.2107e-2 (6.12e-3) $+$       & 1.9836e-1 (1.02e-2) $+$            \\
            \hline
            \multirow{6}{*}{MaF11} & 5   & 2.7952e-2 (8.10e-4)      & 3.2286e-2 (1.55e-3) $+$       & \hl{2.7763e-2 (8.77e-4) $\approx$} & 3.0983e-2 (1.68e-3) $+$       & 5.6892e-2 (5.07e-3) $+$ & 4.0967e-2 (1.32e-3) $+$       & 3.8974e-2 (1.04e-3) $+$            & 7.2990e-2 (1.08e-2) $+$       & 4.9126e-2 (1.63e-3) $+$            \\
                                   & 8   & 2.9233e-2 (7.36e-4)      & 3.3787e-2 (1.83e-3) $+$       & \hl{2.8458e-2 (6.94e-4) $-$}       & 3.3817e-2 (2.74e-3) $+$       & 6.8683e-2 (5.48e-3) $+$ & 4.2883e-2 (1.11e-3) $+$       & 3.7203e-2 (2.05e-3) $+$            & 4.0437e-2 (2.24e-3) $+$       & 8.6705e-2 (4.66e-3) $+$            \\
                                   & 10  & 2.8309e-2 (1.13e-3)      & 3.3339e-2 (1.78e-3) $+$       & \hl{2.7662e-2 (8.73e-4) $-$}       & 3.0797e-2 (2.66e-3) $+$       & 7.4130e-2 (6.46e-3) $+$ & 3.8129e-2 (1.69e-3) $+$       & 3.9213e-2 (3.96e-3) $+$            & 3.7534e-2 (1.27e-3) $+$       & 1.1730e-1 (8.21e-3) $+$            \\
                                   & 13  & 2.7310e-2 (1.50e-3)      & 3.5161e-2 (3.01e-3) $+$       & \hl{2.6858e-2 (1.40e-3) $\approx$} & 5.3050e-2 (1.03e-2) $+$       & 8.6739e-2 (3.35e-2) $+$ & 3.1424e-2 (1.30e-3) $+$       & 2.7052e-2 (1.63e-3) $\approx$      & 3.1735e-2 (2.59e-3) $+$       & 1.4263e-1 (2.75e-2) $+$            \\
                                   & 15  & 2.5593e-2 (1.90e-3)      & 3.0983e-2 (4.35e-3) $+$       & 2.6044e-2 (1.24e-3) $\approx$      & 5.2369e-2 (1.06e-2) $+$       & 8.3706e-2 (1.17e-2) $+$ & 2.6583e-2 (2.00e-3) $\approx$ & \hl{2.5448e-2 (1.42e-3) $\approx$} & 2.6107e-2 (3.67e-3) $\approx$ & 1.2525e-1 (3.37e-2) $+$            \\
                                   & 20  & 2.0319e-2 (1.35e-3)      & 2.8384e-2 (2.63e-3) $+$       & \hl{1.8429e-2 (1.05e-3) $-$}       & 4.8823e-2 (1.14e-2) $+$       & 9.6916e-2 (2.78e-2) $+$ & 2.4227e-2 (1.63e-3) $+$       & 3.0975e-2 (4.51e-3) $+$            & 3.3353e-2 (9.05e-3) $+$       & 8.6101e-2 (3.03e-2) $+$            \\
            \hline
            \multirow{6}{*}{MaF12} & 5   & \hl{8.6568e-2 (1.52e-3)} & 8.8287e-2 (3.29e-3) $+$       & 8.7330e-2 (1.77e-3) $\approx$      & 1.0212e-1 (2.09e-3) $+$       & 2.1941e-1 (3.11e-2) $+$ & 1.0502e-1 (6.48e-3) $+$       & 1.0007e-1 (5.47e-3) $+$            & 1.5542e-1 (2.23e-2) $+$       & 1.2044e-1 (1.42e-2) $+$            \\
                                   & 8   & \hl{1.6295e-1 (2.86e-3)} & 1.7071e-1 (6.85e-3) $+$       & 1.6557e-1 (3.07e-3) $+$            & 2.1666e-1 (1.52e-2) $+$       & 4.5483e-1 (3.80e-2) $+$ & 1.9021e-1 (1.36e-2) $+$       & 1.8436e-1 (9.48e-3) $+$            & 1.9826e-1 (1.51e-2) $+$       & 2.4462e-1 (1.25e-2) $+$            \\
                                   & 10  & \hl{1.9835e-1 (3.07e-3)} & 2.0204e-1 (9.96e-3) $\approx$ & 1.9853e-1 (3.36e-3) $\approx$      & 2.4436e-1 (1.41e-2) $+$       & 4.9065e-1 (3.93e-2) $+$ & 2.1796e-1 (6.59e-3) $+$       & 2.1120e-1 (6.80e-3) $+$            & 2.3987e-1 (1.35e-2) $+$       & 3.0487e-1 (1.54e-2) $+$            \\
                                   & 13  & 2.4694e-1 (6.62e-3)      & 2.4908e-1 (1.20e-2) $\approx$ & \hl{2.4508e-1 (7.67e-3) $\approx$} & 3.2423e-1 (1.67e-2) $+$       & 5.8254e-1 (4.57e-2) $+$ & 2.5836e-1 (1.18e-2) $+$       & 2.6695e-1 (1.23e-2) $+$            & 2.7936e-1 (1.85e-2) $+$       & 3.8020e-1 (1.59e-2) $+$            \\
                                   & 15  & 2.5683e-1 (7.12e-3)      & 2.6301e-1 (1.20e-2) $\approx$ & \hl{2.5473e-1 (7.15e-3) $\approx$} & 2.9995e-1 (2.09e-2) $+$       & 5.8813e-1 (3.88e-2) $+$ & 2.7365e-1 (1.28e-2) $+$       & 2.7247e-1 (1.05e-2) $+$            & 2.9527e-1 (1.64e-2) $+$       & 3.9282e-1 (1.38e-2) $+$            \\
                                   & 20  & 3.1165e-1 (9.06e-3)      & 2.8049e-1 (1.65e-2) $-$       & 3.0784e-1 (8.20e-3) $-$            & 4.5069e-1 (2.99e-2) $+$       & 7.3029e-1 (3.35e-2) $+$ & 3.5999e-1 (1.03e-2) $+$       & \hl{2.8003e-1 (1.06e-2) $-$}       & 3.8759e-1 (3.14e-2) $+$       & 5.0637e-1 (1.87e-2) $+$            \\
            \hline
            \multirow{6}{*}{MaF13} & 5   & \hl{5.3461e-2 (8.79e-3)} & 2.2117e-1 (4.64e-2) $+$       & 5.9295e-2 (8.91e-3) $+$            & 1.1750e-1 (1.82e-2) $+$       & 3.3875e-1 (1.32e-2) $+$ & 9.4261e-2 (1.31e-2) $+$       & 7.4144e-2 (6.68e-3) $+$            & 1.4097e-1 (9.35e-3) $+$       & 6.0467e-2 (8.47e-3) $+$            \\
                                   & 8   & \hl{5.2642e-2 (5.97e-3)} & 4.0831e-1 (4.07e-1) $+$       & 5.9268e-2 (1.02e-2) $+$            & 1.2718e-1 (2.06e-2) $+$       & 3.9145e-1 (8.50e-3) $+$ & 9.9311e-2 (1.40e-2) $+$       & 8.0485e-2 (5.39e-3) $+$            & 2.0193e-1 (2.83e-2) $+$       & 5.5665e-2 (7.00e-3) $\approx$      \\
                                   & 10  & \hl{4.9611e-2 (7.13e-3)} & 3.8016e-1 (5.41e-2) $+$       & 5.7837e-2 (9.84e-3) $+$            & 1.2753e-1 (1.78e-2) $+$       & 4.1216e-1 (1.36e-2) $+$ & 9.0345e-2 (1.34e-2) $+$       & 7.9873e-2 (4.55e-3) $+$            & 1.8253e-1 (1.56e-2) $+$       & 5.1915e-2 (6.23e-3) $\approx$      \\
                                   & 13  & 5.6098e-2 (5.25e-3)      & 4.8650e-1 (4.79e-1) $+$       & 5.5433e-2 (9.42e-3) $\approx$      & 1.7028e-1 (4.14e-2) $+$       & 4.5222e-1 (5.98e-3) $+$ & 1.2665e-1 (2.28e-2) $+$       & 1.0196e-1 (8.34e-3) $+$            & 1.6457e-1 (2.00e-2) $+$       & \hl{5.2647e-2 (3.44e-3) $-$}       \\
                                   & 15  & \hl{4.7227e-2 (3.83e-3)} & 4.1153e-1 (4.95e-2) $+$       & 5.1783e-2 (8.45e-3) $+$            & 1.5840e-1 (7.40e-2) $+$       & 4.6594e-1 (1.37e-2) $+$ & 1.0655e-1 (1.56e-2) $+$       & 9.3031e-2 (5.54e-3) $+$            & 1.5770e-1 (2.15e-2) $+$       & 4.7716e-2 (3.53e-3) $\approx$      \\
                                   & 20  & 4.7404e-2 (3.41e-3)      & 4.2420e-1 (3.33e-2) $+$       & 4.8868e-2 (4.21e-3) $\approx$      & 1.5905e-1 (3.76e-2) $+$       & 5.0777e-1 (2.16e-3) $+$ & 1.1030e-1 (1.82e-2) $+$       & 1.0511e-1 (7.48e-3) $+$            & 1.6619e-1 (1.85e-2) $+$       & \hl{4.6314e-2 (2.85e-3) $\approx$} \\
            \bottomrule
        \end{tabular}
    }
    \\
	\vspace{1mm}
	\footnotesize \raggedright \hspace{4mm}Each column represents each algorithm, and each row represents each test instance.
	
	\hspace{4mm}Each cell shows the median and standard deviation of the evaluations of HV and IGD${}^+$ in 31 runs.
	
	\hspace{4mm}The algorithm with the best median value for each row is highlighted.
	
	\hspace{4mm}The result of comparison with RVEA-CA at a 0.05 level by Wilcoxon's rank-sum test is shown on the right end of each cell in each row except for RVEA-CA.
	
	\hspace{4mm}``$+$'', ``$\approx$'', and ``$-$'' mean that RVEA-CA performs significantly better than, non-significantly better/worse than, and significantly worse than each algorithm, respectively.
    \label{tbl:IGDP2}
\end{sidewaystable*}

\begin{table*}[htbp]
    \centering
    \caption{Summary of comparisons with the significance of the difference to RVEA-CA}
    \label{tbl:summaryOfResult}
    \begin{tabular}{cccccccccc}
        \toprule
        \multirow{2}{*}{Property}     & \multirow{2}{*}{Indicator} & \multicolumn{8}{c}{RVEA-CA vs. ($+$/$\approx$/$-$)}                                                                                                                                                                   \\ \cmidrule(){3-10}
                                      &                            & RVEA                                                & RVEA-iGNG             & DEA-GNG              & A-NSGA-III           & VaEA                 & AR-MOEA              & MOEA/D-AWA           & AdaW                 \\
        \midrule
        \multirow{2}{*}{Overall}      & HV                         & \cellcolor{hl}65/5/8                                              & \cellcolor{hl}40/14/24              & \cellcolor{hl}55/12/11             & \cellcolor{hl}73/3/2               & \cellcolor{hl}42/8/28              & \cellcolor{hl}40/5/33              & \cellcolor{hl}53/7/18              & \cellcolor{hl}43/9/26              \\
                                      & IGD${}^+$                  & \cellcolor{hl}58/6/14                                             & \cellcolor{hl}30/24/24              & \cellcolor{hl}48/16/14             & \cellcolor{hl}73/2/3               & \cellcolor{hl}49/5/24              & \cellcolor{hl}39/6/33              & \cellcolor{hl}63/2/13              & \cellcolor{hl}45/11/22             \\ \midrule
        \multirow{2}{*}{Linear}       & HV                         & \cellcolor{hl}18/0/0                                & 5/3/10                & \cellcolor{hl}10/3/5 & \cellcolor{hl}18/0/0 & 4/0/14               & 4/1/13               & \cellcolor{hl}16/2/0 & 8/2/8                \\
                                      & IGD${}^+$                  & \cellcolor{hl}18/0/0                                & 6/6/6                 & \cellcolor{hl}9/3/6  & \cellcolor{hl}18/0/0 & 7/0/11               & 5/0/13               & \cellcolor{hl}18/0/0 & 8/2/8                \\ \cmidrule{1-10}
        \multirow{2}{*}{Concave}      & HV                         & \cellcolor{hl}25/0/5                                & \cellcolor{hl}15/6/9  & \cellcolor{hl}20/7/3 & \cellcolor{hl}27/2/1 & \cellcolor{hl}19/6/5 & \cellcolor{hl}23/2/5 & \cellcolor{hl}23/2/5 & \cellcolor{hl}20/3/7 \\
                                      & IGD${}^+$                  & \cellcolor{hl}21/3/6                                & \cellcolor{hl}11/12/7 & \cellcolor{hl}19/9/2 & \cellcolor{hl}27/1/2 & \cellcolor{hl}22/3/5 & \cellcolor{hl}23/0/7 & \cellcolor{hl}24/0/6 & \cellcolor{hl}14/7/9 \\ \cmidrule{1-10}
        \multirow{2}{*}{Convex}       & HV                         & \cellcolor{hl}16/2/0                                & \textbf{15/1/2}       & \cellcolor{hl}14/2/2 & \cellcolor{hl}17/1/0 & \cellcolor{hl}16/2/0 & 8/2/8                & \cellcolor{hl}9/3/6  & \cellcolor{hl}10/4/4 \\
                                      & IGD${}^+$                  & \cellcolor{hl}11/2/5                                & 7/4/7                 & \cellcolor{hl}14/2/2 & \cellcolor{hl}17/1/0 & \cellcolor{hl}16/2/0 & 4/5/9                & \cellcolor{hl}12/2/4 & \cellcolor{hl}12/2/4 \\ \cmidrule{1-10}
        \multirow{2}{*}{Mixed}        & HV                         & \cellcolor{hl}6/3/3                                 & \cellcolor{hl}5/4/3   & \cellcolor{hl}11/0/1 & \cellcolor{hl}11/0/1 & 3/0/9                & 5/0/7                & 5/0/7                & 5/0/7                \\
                                      & IGD${}^+$                  & \cellcolor{hl}8/1/3                                 & \cellcolor{hl}6/2/4   & \cellcolor{hl}6/2/4  & \cellcolor{hl}11/0/1 & 4/0/8                & \textbf{7/1/4}       & \textbf{9/0/3}       & \textbf{11/0/1}      \\ \cmidrule{1-10}
        \multirow{2}{*}{Degenerate}   & HV                         & \cellcolor{hl}19/0/5                                & 6/5/13                & \cellcolor{hl}16/6/2 & \cellcolor{hl}21/2/1 & 10/2/12              & 9/2/13               & \cellcolor{hl}17/2/5 & \cellcolor{hl}15/2/7 \\
                                      & IGD${}^+$                  & \cellcolor{hl}19/0/5                                & 5/9/10                & \cellcolor{hl}16/7/1 & \cellcolor{hl}21/1/2 & \textbf{14/1/9}      & 11/0/13              & \cellcolor{hl}19/0/5 & \cellcolor{hl}10/6/8 \\ \cmidrule{1-10}
        \multirow{2}{*}{Disconnected} & HV                         & \cellcolor{hl}11/0/1                                & \cellcolor{hl}10/1/1  & \cellcolor{hl}11/0/1 & \cellcolor{hl}11/0/1 & \cellcolor{hl}9/0/3  & \cellcolor{hl}9/1/2  & \cellcolor{hl}6/2/4  & \cellcolor{hl}9/1/2  \\
                                      & IGD${}^+$                  & \cellcolor{hl}11/0/1                                & \cellcolor{hl}5/3/4   & \cellcolor{hl}10/1/1 & \cellcolor{hl}11/0/1 & \cellcolor{hl}7/1/4  & \cellcolor{hl}9/2/1  & \cellcolor{hl}10/1/1 & \cellcolor{hl}11/0/1 \\ \cmidrule{1-10}
        \multirow{2}{*}{Biased}       & HV                         & \cellcolor{hl}7/3/2                                 & \cellcolor{hl}7/3/2   & \textbf{9/1/2}       & \cellcolor{hl}12/0/0 & 5/1/6                & 1/0/11               & 6/0/6                & 3/2/7                \\
                                      & IGD${}^+$                  & \cellcolor{hl}7/1/4                                 & \cellcolor{hl}7/2/3   & 5/2/5                & \cellcolor{hl}12/0/0 & \textbf{8/0/4}       & 2/1/9                & \textbf{10/0/2}      & \textbf{9/2/1}       \\ \cmidrule{1-10}
        \multirow{2}{*}{Multimodal}   & HV                         & \cellcolor{hl}15/2/1                                & \cellcolor{hl}12/2/4  & \cellcolor{hl}13/3/2 & \cellcolor{hl}16/1/1 & \cellcolor{hl}11/3/4 & \cellcolor{hl}13/2/3 & \cellcolor{hl}13/1/4 & \cellcolor{hl}9/4/5  \\
                                      & IGD${}^+$                  & \cellcolor{hl}12/2/4                                & \cellcolor{hl}11/1/6  & \cellcolor{hl}10/6/2 & \cellcolor{hl}16/1/1 & \cellcolor{hl}9/3/6  & \cellcolor{hl}10/3/5 & \cellcolor{hl}11/1/6 & \cellcolor{hl}8/3/7  \\ \cmidrule{1-10}
        \multirow{2}{*}{Nonseparable} & HV                         & \cellcolor{hl}18/0/0                                & \cellcolor{hl}15/3/0  & \cellcolor{hl}16/1/1 & \cellcolor{hl}18/0/0 & \cellcolor{hl}15/2/1 & \cellcolor{hl}16/1/1 & \cellcolor{hl}11/4/3 & \cellcolor{hl}16/1/1 \\
                                      & IGD${}^+$                  & \cellcolor{hl}14/3/1                                & \cellcolor{hl}5/9/4   & \cellcolor{hl}18/0/0 & \cellcolor{hl}18/0/0 & \cellcolor{hl}17/1/0 & \cellcolor{hl}15/2/1 & \cellcolor{hl}17/1/0 & \cellcolor{hl}13/4/1 \\
        \bottomrule
    \end{tabular}
    \\
	\vspace{1mm}
	\footnotesize \raggedright \hspace{4mm}``$+$'', ``$\approx$'', and ``$-$'' mean the number of test instances that RVEA-CA performs significantly better than, non-significantly better/worse than, and
	
	\hspace{4mm}significantly worse than each algorithm, respectively.
	
	\hspace{4mm}If RVEA-CA outperforms its competitor in terms of both HV and IGD${}^+$, the cell is highlighted
	
	\hspace{4mm}If RVEA-CA outperforms its competitor in terms of either HV or IGD${}^+$, the result is emphasized in bold.
\end{table*}

The results for all properties (the row denoted ``Overall'') in Table \ref{tbl:summaryOfResult} show that RVEA-CA has better performance than the other algorithms under comparison in terms of both HV and IGD${}^+$.
Therefore, RVEA-CA can be regarded as having comprehensively higher search performance than other algorithms.
In addition, these results also show that RVEA-CA outperforms A-NSGA-III on almost all test instances and outperforms RVEA, DEA-GNG, and MOEA/D-AWA on many test instances, while RVEA-iGNG, VaEA, AR-MOEA, and AdaW have competitive performance with RVEA-CA.

The results in Table \ref{tbl:summaryOfResult} show that RVEA-CA outperforms RVEA-iGNG on Concave, Convex, Mixed, Disconnected, Biased, Multimodal, and Nonseparable properties in terms of HV and outperforms RVEA-iGNG on Concave, Mixed, Disconnected, Biased, Multimodal, Nonseparable properties in terms of IGD${}^+$.
These results show the effectiveness of the proposed mechanisms introduced in RVEA-CA, i.e., mating selection strategy based on clustering, employing CA instead of GNG, and automatic parameter adjusting.

CA has only two hyperparameters to be adjusted, while GNG has six.
Therefore, we can easily handle CA, and RVEA-CA achieves obtaining an appropriate network by automatic parameter adjusting.
On the other hand, some issues remain, especially RVEA-CA is outperformed by RVEA-iGNG on Degenerate property in terms of both HV and IGD${}^+$.
This may be caused by the characteristics of CA.
CA creates nodes directly at highly novel input locations, whereas GNG creates nodes at mid-position between nodes that are frequently updated.
Therefore, the nodes in CA are evenly distributed for the input distribution, while the nodes in GNG are distributed in the middle among the inputs.
As a result of degenerated problems, a wider CA network causes weak convergence of RVEA-CA.

The results in Table \ref{tbl:summaryOfResult} show that the number of test instances where RVEA-CA outperforms VaEA is more than twice the number of test instances where RVEA-CA is inferior to VaEA in terms of IGD${}^+$.
Table \ref{tbl:summaryOfResult} also shows that RVEA-CA outperforms VaEA on Concave, Convex, Disconnected, Multimodal, and Nonseparable properties in terms of HV and outperforms VaEA on Concave, Convex, Degenerate, Disconnected, Biased, Multimodal, and Nonseparable properties in terms of IGD${}^+$.
Since VaEA is a dominance-based MOEA, which is generally considered to have poor search performance in MaOP, it is surprising that VaEA shows results competitive with RVEA-CA and implies the potential of domaince-based MOEAs.
Especially, the results on Mixed property may show the validity of dominance relation in irregular MaOPs.

The results in Table \ref{tbl:summaryOfResult} show that AR-MOEA is the most competitive with RVEA-CA among the eight algorithms.
Table \ref{tbl:summaryOfResult} also shows that RVEA-CA outperforms AR-MOEA on Concave, Disconnected, Multimodal, and Nonseparable properties in terms of HV and outperforms AR-MOEA on Concave, Mixed, Disconnected, Multimodal, and Nonseparable properties in terms of IGD${}^+$.
These results show that RVEA-CA has some issues against indicator-based MOEAs and indicator-based MOEAs are promising.

The results in Table \ref{tbl:summaryOfResult} show that the number of test instances where RVEA-CA outperforms AdaW is more than twice the number of test instances where RVEA-CA is inferior to AdaW in terms of IGD${}^+$.
Table \ref{tbl:summaryOfResult} also shows that RVEA-CA outperforms AdaW on Concave, Convex, Degenerate, Disconnected, Multimodal, and Nonseparable properties in terms of HV and outperforms AdaW on Concave, Convex, Mixed, Degenerate, Disconnected, Biased, Multimodal, and Nonseparable properties in terms of IGD${}^+$.
AdaW is an adaptive decomposition-based MOEA, but does not use the clustering-based approach.
Therefore, these results show the effectiveness of the clustering-based approach.
Especially, the results on Degenerate and Disconnected properties show that the clustering-based approach is effective in irregular MaOPs.

\subsection{Discussion}

In this paper, we compare the search performance of algorithms in various types of test problems, and Table \ref{tbl:summaryOfResult} summarizes the results of comparisons for each property of test problems.
Considering the experimental results for each property of the problems, we can see the advantages and disadvantages and the issues of RVEA-CA against other algorithms.

On the Disconnected property, Table \ref{tbl:summaryOfResult} shows that RVEA-CA outperforms all other problems in terms of both HV and IGD${}^+$.
This is an obvious advantage of RVEA-CA against other algorithms and shows the effectiveness of the mating strategy based on clustering.
In addition, RVEA-CA also outperforms all other algorithms on Concave property and outperforms almost all algorithms on Convex property.
These results show the advantage of RVEA-CA not only in irregular problems but also in regular problems.

On the other hand, Table \ref{tbl:summaryOfResult} shows the disadvantage of RVEA-CA:
RVEA-CA is outperformed by RVEA-iGNG, VaEA, and AR-MOEA on Linear property, by VaEA on Mixed property, and by RVEA-iGNG and AR-MOEA on Degenerate property, and by AR-MOEA on Biased property.
In particular, since Linear is one of the regular properties of MOPs and we can not figure out the cause behind these results, this disadvantage is a major issue.

Moreover, the results on Biased property imply the issue of adaptive decomposition-based MOEAs employing the clustering-based approach.
While RVEA-CA nearly outperforms RVEA-iGNG and DEA-GNG on Biased property, AR-MOEA outperforms RVEA-CA, and VaEA and AdaW are competitive with RVEA-CA on Biased property.
This may be due to the failure of ideal point estimation and normalization of the objective function caused by intensive search in local regions using a clustering-based approach.

\begin{figure}[htb]
    \centering
    \includegraphics[width=200pt]{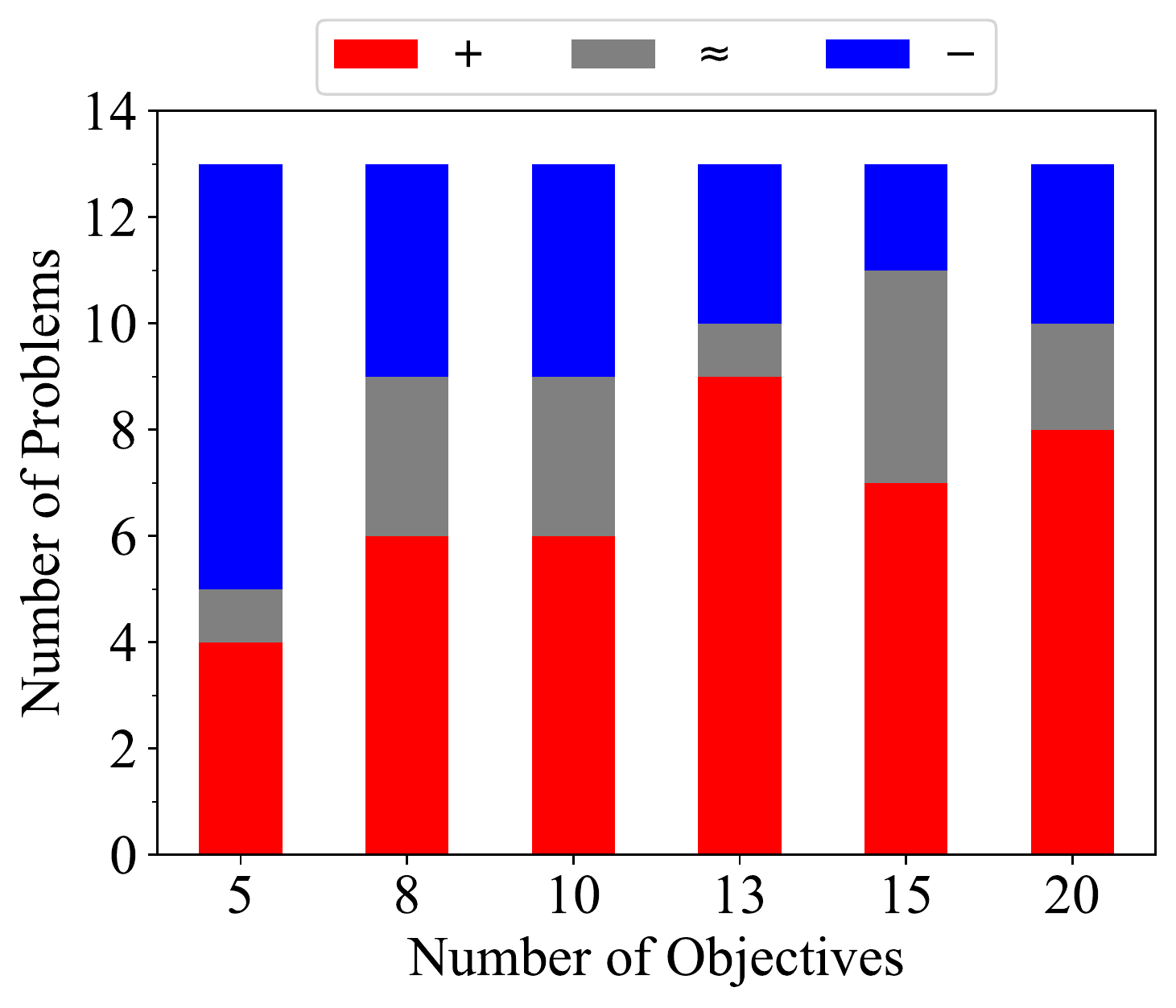}
    \caption{Aggregated results of RVEA-CA against RVEA-iGNG for each number of objective}
    \label{fig:objectivewise}
\end{figure}

In this paper, we compare the search performance of algorithms in 5-, 8-, 10-, 15-, and 20-objective test problems.
Considering the experimental results for each number of objectives, we can see a trend in the relative change in search performance for changes in the number of objectives.
Fig. \ref{fig:objectivewise} aggregates the results of comparison from Tables \ref{tbl:HV1}-\ref{tbl:HV2} between RVEA-CA and RVEA-iGNG for each number of objectives.
Fig. \ref{fig:objectivewise} shows that as the number of objectives increases, the number of problems in which RVEA-CA outperforms RVEA-iGNG increases.
Tables \ref{tbl:HV1}-\ref{tbl:HV2} show that the comparison results for the MaF7 problem show a similar trend, and Table \ref{tbl:MaF} shows that the MaF7 problem has Disconnected property.
From these results, the basic idea of RVEA-CA, intensive local search by more actively utilizing information on the distribution of solutions obtained during the search process, is considered to be more effective for irregular MaOPs, where convergence to a local region is more important, as the number of objectives increases.

\section{Conclusion}\label{sec:conclusion}

In this paper, we proposed RVEA-CA for various types of MaOPs.
RVEA-CA adjusts the distribution of the reference vector set in RVEA by learning the distribution of solutions obtained during the search process with CA.
RVEA-CA also adjusts the parameters of CA in the search process to obtain an appropriate network from CA.
Moreover, we proposed a new mating selection strategy based on clustering as a method to utilize network information more actively and efficiently than existing MOEAs.

We compared RVEA-CA with RVEA, RVEA-iGNG, DEA-GNG, A-NSGA-III, VaEA, AR-MOEA, MOEA/D-AWA, and AdaW by applying them to 78 test problems.
The experimental results showed that RVEA-CA has higher search performance on various types of MaOPs than existing algorithms in terms of both HV and IGD${}^+$.
In addition, we investigated the effects of properties and the number of objectives of test problems on the search performance of REVA-CA.

We will focus on developing a method for analyzing search results in MaOPs as one of our future works.
One possible approach is mapping the population into feature space using some modified indicators from high-dimensional objective space.
Another future direction is to solve some issues found in this paper. 
In particular, a promising direction is the development of a mechanism to improve the accuracy of the ideal point estimation.

\bibliographystyle{ieeetr}
\bibliography{myref}

\EOD

\end{document}